%% file: main.tex
\documentclass[conference]{IEEEtran}
\makeatletter
\def\endthebibliography{%
	\def\@noitemerr{\@latex@warning{Empty `thebibliography' environment}}%
	\endlist
}
\makeatother
\usepackage{cite}
\usepackage{graphicx}
\usepackage{textcomp}
\usepackage{xcolor}
\def\BibTeX{{\rm B\kern-.05em{\sc i\kern-.025em b}\kern-.08em
    T\kern-.1667em\lower.7ex\hbox{E}\kern-.125emX}}
    
\usepackage{amssymb}		
\usepackage{amsthm,amsmath,amsfonts}
\usepackage{graphicx}
\usepackage{algorithm,algpseudocode} 
\usepackage{breakcites}
\usepackage{float}
\usepackage{bbm}
\usepackage{bm}
\usepackage{extarrows}
\usepackage{enumitem}
\usepackage{comment}
\usepackage{mathrsfs}
\usepackage{subcaption}
\usepackage{booktabs}
\usepackage[switch]{lineno}

\newtheorem{theorem}{Theorem}
\newtheorem{lemma}[theorem]{Lemma}

\newtheorem{proposition}[theorem]{Proposition}
\newtheorem{definition}{Definition}
\newtheorem{assumption}{Assumption}
\theoremstyle{remark}
\newtheorem{remark}{Remark} 

\usepackage{xspace}
\newcommand{\R}{\mathbb{R}}

\newcommand{\dis}{\displaystyle}
\newcommand{\E}{\mathbb{E}\,}

\newcommand{\x}{\mathbf{x}}
\newcommand{\y}{\mathbf{y}}
\newcommand{\z}{\mathbf{z}}

\newcommand{\e}{\mathrm{e}}
\newcommand{\<}{\langle}
\renewcommand{\>}{\rangle}

\newcommand{\eps}{\varepsilon}
\newcommand{\noise}{\boldsymbol{\epsilon}}

\usepackage{tikz}

\newcommand{\transpose}{\intercal} 
\newcommand{\trans}{\transpose}

\newcommand{\gradsvrg}{\widetilde{\nabla}}
\newcommand{\xbatch}{\widetilde{\x}}
\newcommand{\gradnorm}{\|\nabla _t\|^2}
\newcommand{\fsp}{\x_{\mathrm{fsp}}}
\newcommand{\pr}{\mathbf{Pr}}

\begin{document}
\title{Stochastic Gradient Langevin Dynamics with Variance Reduction}

\author{\IEEEauthorblockN{ Zhishen Huang}
\IEEEauthorblockA{\textit{Department of Computational Mathematics}\\
\textit{Michigan State University}\\
East Lansing, MI, USA \\
zhishen.huang@colorado.edu}
\and
\IEEEauthorblockN{ Stephen Becker}
\IEEEauthorblockA{\textit{Department of Applied Mathematics} \\
\textit{University of Colorado Boulder}\\
Boulder, CO, USA \\
stephen.becker@colorado.edu}

}

\maketitle
\begin{abstract}
Stochastic gradient Langevin dynamics (SGLD) has gained the attention of optimization researchers due to its global optimization properties. This paper proves an improved convergence property to local minimizers of nonconvex objective functions using SGLD accelerated by variance reductions. Moreover, we prove an ergodicity property of the SGLD scheme, which gives insights on its potential to find global minimizers of nonconvex objectives.

\end{abstract}

\section{Introduction}
In this paper we consider the optimization algorithm \textit{stochastic gradient descent} (SGD) with variance reduction (VR) and Gaussian noise injected at every iteration step. For historical reasons, the particular randomization format of injecting Gaussian noises bears the name \textit{Langevin dynamics} (LD). Thus, the scheme we consider is referred as \textit{stochastic gradient Langevin dynamics with variance reduction} (SGLD-VR). We prove the ergodicity property of SGLD-VR schemes when used as an optimization algorithm, which the normal SGD method without the additional noise does not have. As the ergodicity property implies the non-trivial probability for the LD process to visit the whole space, the set of global minima will also be traversed during the iteration. 
We also provide convergence results of SGLD-VR to local minima in a similar style to \cite{SGLD_Gu}. Taken together, the results show that SGLD-VR concentrates around local minima, but is never stuck at a particular point, and thus is useful for global optimization.

\newcommand{\data}{\xi} 
We apply the SGLD-VR scheme on the empirical risk minimization (ERM) problem:
\begin{equation}
    \label{prob::LD_main_prob}
    \textrm{minimize } f(\boldsymbol{\omega}) = \frac{1}{n} \sum_{i=1}^n f_i(\boldsymbol{\omega})
\end{equation}
which often arises as a sampled version of the stochastic optimization problem: $\dis \min_{\boldsymbol{\omega}} F(\boldsymbol{\omega}) = \int f(\boldsymbol{\omega};\data)\,\mathrm{d} \data$, where $\data$ is the collection of training data and $\boldsymbol{\omega}$ is the parameter for the model, i.e., $F(\boldsymbol{\omega})$ may be the expectation of a loss function with respect to stochastic data $\data$,
 $F(\boldsymbol{\omega}) = \E \ell( \boldsymbol{\omega}, \data)$, so then $f_i(\boldsymbol{\omega}) = \ell(\boldsymbol{\omega}, \data_i )$ for i.i.d.\ realizations $(\data_i)_{i=1}^n$. In the following text we use $\x$ instead of $\boldsymbol{\omega}$ as the input for the objective $f$ in order to conform to optimization literature conventions. 
 The usual SGD framework for ERM problems is at every gradient step to form a minibatch subsampled from $\{1,\ldots,n\}$ and include only the sampled terms (with a reweighting) in the sum, in order to reduce computational complexity.
 %
 Variance reduction which exploits the finite-sum structure of Eq.~\eqref{prob::LD_main_prob}  accelerates convergence, and LD is is used in the SGD scheme to enable the ergodicity property. 

\subsection{Prior art}

The Langevin dynamical equation describes the trajectory $X(t)$ of the following stochastic differential equation
\begin{equation}
\mathrm{d}X_t = -\nabla U(X_t)\,\mathrm{d} t + \sigma\,\mathrm{d}B_t,
\end{equation}
where $B_t$ is a Wiener process. This equation characterizes the continuous motion of a particle subject to fluctuations (due to nonzero temperature) in a potential field $U$,
and in the limit $\sigma\to\infty$ approaches Brownian motion. Using this dynamic as a master equation, through Kramers–Moyal expansion one can derive the Fokker-Planck equation, which gives the spatial distribution of particles at a given time, thus a full characterization of the statistical properties of a particle ensemble~\cite{KRAMERS_FK_equation}. 

\paragraph{LD and sampling} The connection between Langevin dynamics (LD) and the distribution of particle ensemble reveals the potential of applying LD on sampling. Suppose that the  distribution of interest is $\pi(\x)$ and that there exists a function $U$ such that $\pi(\x)=\frac{\exp(-U(\x))}{\int\exp(-U(\x))\,\mathrm{d}\x} $, then the LD equation 
using this $U$ defines a stochastic process with stationary distribution $\pi(\x)$.
To numerically implement the LD equation for sampling purposes, one needs to discretize the continuous LD equation. A simple version of the discretization is the \textit{unadjusted Langevin algorithm} (ULA), 
\begin{equation}
    \x_{k+1} = \x_k + \Delta \x_k, \;
    \Delta \x_k = - \eta_k\nabla U(\x_k) + \rho_0\sqrt{\eta_k}
    \noise_k
\end{equation}
where $\noise_k \sim \mathcal{N}(0,\mathbf{I}_d)$, $\x\in \R^d$
and $\rho_0=\sigma$.
The Gaussian noise term enables the scheme to explore the sample space and the drift term guides the direction of exploration. One common modified scheme is the \textit{Metropolis adjusted Langevin algorithm} (MALA), where upon the suggested update by ULA, there is an additional accept/reject step, with the probability of accepting the the update as $1\wedge\frac{\pi(\x_k)p(\x_k|\x_{k+1})}{\pi(\x_{k+1}) p(\x_{k+1}|\x_k) }$. 

Naturally two central questions related to this sample scheme arise: whether or not the distribution of samples generated by LD converges, and if so, to $\pi$; and what is the mixing time of LD (i.e., how long does it takes for the LD to approximately reach equilibrium hence generating valid samples from the distribution $\pi$).
The first question motivates the importance of MALA: in terms of the \textit{total variation} (TV) distance, while ULA can fail to converge for either light-tailed or heavy-tailed target distribution, MALA is guaranteed to converge to any continuous target distribution \cite{MeynTweedie_MC_Book}. Regarding the second question about convergence speed, researchers have investigated the sufficient conditions for ULA and MALA respectively to guarantee exponential (geometric) convergence to target distribution. \cite{MengersenTweedie_MHA} show for distributions over $\R$, the necessary and sufficient condition for MALA to converge to target distribution $\pi(\x)$ at geometric speed is that $\pi(\x)$ has exponential tails. The sufficiency of this condition is generalized to higher dimension in \cite{RobertsTweedie_MHA}.
The seminal work by \cite{RobertsTweedie_LD} shows that
MALA cannot converge at geometric speed to target distributions that are in essence non-localized, or heavy-tailed.
%

In parallel there have been works to show the convergence of LD for distribution approximation in terms of Wasserstein-2 distance \cite{dalalyan_LMC_guarantees} and KL-divergence respectively \cite{LMC_convergence_KL}.

A particular case of interest for the application of LD on sampling is to find the posterior distribution of parameters 
$\x$
in the Bayesian setting, where the updates are set as
\begin{equation}
\label{eqn::bayesian_theta_update}
\Delta \x_k = \eta_k \left( \nabla p(\x_k)  + \sum_{i=1}^N \nabla \log p(\data_i \mid \x_k) \right) + \sqrt{\eta_k}\noise_k
\end{equation}
where $\noise_k\sim\mathcal{N}(0,\mathbf{I})$
and $p(\x,\data)$ is the joint probability of parameters $\x$ and data $\data$. To maximize the likelihood, \cite{Bayes_learning_SGLD} suggest to use the format of stochastic gradient descent in the derivative term of \eqref{eqn::bayesian_theta_update}. \cite{Borkar_Mitter_1999} show that this minibatch-styled LD will converge to the correct distribution in terms of KL divergence.

\paragraph{LD and optimization} The main focus of this paper is on optimization. LD offers an exciting opportunity for  global optimization due to the exploring nature of the Brownian motion term. Simulating multiple particles to obtain information about the geometric landscape of the objective function---thus locating a global minimum---is often too computationally expensive to be practical. Notice that when one considers the convergence to a distribution, the exploring nature of LD due to continually injected Gaussian noise of constant variance is the key factor, while for the purpose of optimization, one usually exploits noises with diminishing variance since the goal is to converge to a point. 

The technique to achieve this point convergence is \textit{annealing}, which means decreasing the variance of the noise as $t$ grows. Formally, let the objective function be $U$ and we construct the probability distribution $p_T(\x) = \frac{1}{Z}\exp\left(-\frac{U(\x)}{T} \right)$, where $Z$ is the normalization factor. The key observation is that as the parameter $T\to 0$, the distribution $p_T(\x)$ will concentrate on the global minima. This parameter $T$ is usually referred to as \textit{temperature}, alluding to the alloy annealing process where as temperature decreases, the structure of the metal evolves into the most stable one, hence reaching the state with minimum potential energy. To formulate  LD for optimization, one essentially takes the usual LD equation but with the variance term $\sigma$ now a function of time, $\sigma = \sqrt{T}(t)$:
\[
\mathrm{d}\x_t = \nabla U(\x_t)\,\mathrm{d} t + \sqrt{T(t)}\, \mathrm{d} B_t
\]
The pioneering work by \cite{diffusion_global_optimization_1987} shows that with the annealing schedule $T(t) \propto (\log t)^{-1}$, then  LD will find the global minimum. The work by Chiang et al.\ does not specify how to simulate the continuous version of Langevin dynamics, thus not providing information on the convergence of discrete approximations (such as the Euler-Maruyama method) for LD.  \cite{Gelfand_Mitter_1999} fill this gap by proving that with an annealing schedule $\eta_k \propto k^{-1}$ and $T_k\propto (k\log\log k)^{-1}$, the discretized LD will converge to the global minima in probability, though the convergence may be slow (and improving slow convergence is the motivation for the variance reduction scheme we analyze).
More recently, \cite{SGLD_Raginsky} use optimal transport formalism to study the empirical risk minimization problem. Their proof uses the Wasserstein-2 distance to evaluate distribution discrepancy and consists of two parts: first they show that the discretization error of LD from continuous LD accumulates linearly with respect to the error tolerance level, and then they show that the continuous LD will converge to the true target distribution exponentially fast.

\paragraph{Variance reduction (VR) and LD} In this paper we aim to apply variance reduction techniques in the setting of LD to accelerate the optimization process and to derive an improved time complexity dependence on error tolerance level. 
We use the term ``time complexity'' to be proportional to the iteration count.

The main algorithm we consider in this paper is stochastic gradient Langevin dynamics (SGLD) with variance reduction, which consists of two sources of randomness: one from stochastic gradients, the other from Gaussian noise injected at each step. Previous work have investigated the SGLD for optimization to find local minimizers \cite{Hitting_time_analysis_SGLD_ChenDuTong,Hitting_time_analysis_SGLG_ZhangLiangCharikar}, and reported results for convergence to approximate second order stationary points. 

\begin{figure}
    \centering
    \includegraphics[width=.45\textwidth]{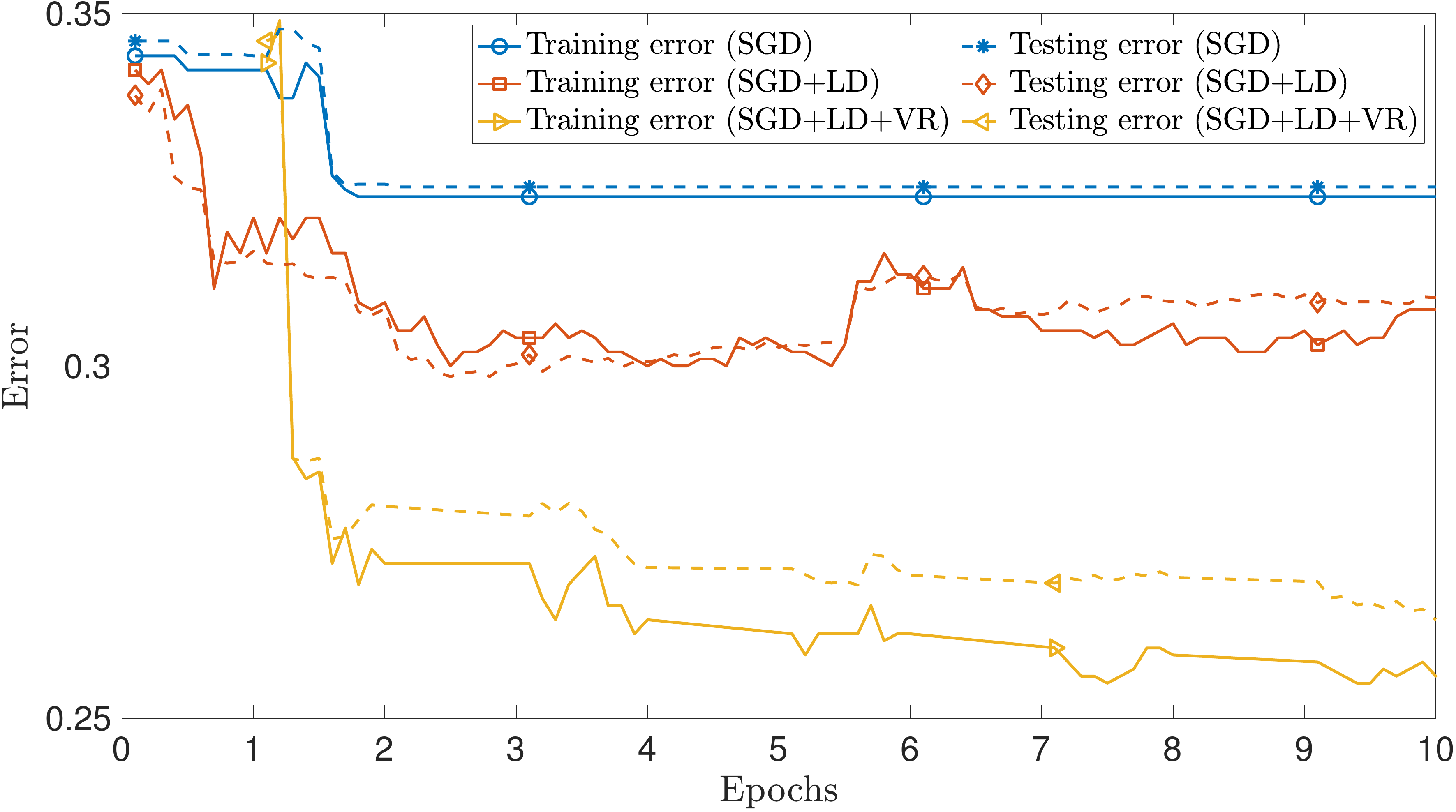}
    \caption{\small 
Example of training a neural net for binary classification with 2 hidden layers, $n=1000$ training points, sigmoid activation function, $\eta_0=10^3$, $\nu=1$, $\rho_0=10^{-2}$, batch size $B_b=100$, and $B_e = 10$. SGLD-VR converges to a good solution, in terms of both training and testing error, more quickly than either SGLD or regular SGD. The $\ell_2^2$ loss was used for training, but the error reported in the figure is the misclassification rate.  }
    \label{fig:1}
\end{figure}

In particular, \cite{SGLD_Gu} show that with constant-variance Gaussian noise injected at each step, SGLD-VR finds an approximate minimizer with time complexity $\mathcal{O}(\frac{\sqrt{n}}{\eps^\frac{5}{2}})\cdot \mathrm{e}^{\mathcal{O}(d)}$, in contrast to the time complexity $\mathcal{O}(\frac{1}{\eps^5}) \cdot \mathrm{e}^{\mathcal{O}(d)} $ for SGLD without variance reduction acceleration, where $n$ is the number of component functions in the ERM; see Figure~\ref{fig:1} which lends some experimental evidence that SGLD-VR can outperform regular SGLD. We aim to improve the dependency on $\eps$ in the analysis. 
We also point out that when the variance of the Gaussian noise is set as constant in SGLD, the function value or the point distance between an optimal point and the iterate can never go to zero, but can at best be bounded by a constant depending on the size of variance.  

The specific VR technique we use was originally proposed to reduce the variance of the minibatch gradient estimator in stochastic gradient descent for convex objectives by \cite{SVRG} and separately by \cite{SAGA}. \cite{Reddi_SVRG_first_order_2016,allen2016variance} have respectively generalized the application of VR techniques to nonconvex objectives and provided convergence guarantee to first-order stationary points.

In essence, we use the \textit{control variate} technique to construct a new gradient estimator by adding an additional term to the minibatch gradient estimator in SGD ($\nabla_{\mathrm{SGD}}$) and this term is correlated with $\nabla_{\mathrm{SGD}}$, thus reducing the variance of the gradient estimator as a whole. More specifically, consider the classic gradient estimator of function $f$ in \eqref{prob::LD_main_prob} at point $\x$: $\nabla_{\mathrm{SGD}} = \frac{1}{|\mathcal{I}|}\sum_{i\in\mathcal{I}} \nabla f_i(\x)$, where $\mathcal{I}\subseteq[n]$. If there is another random variable (r.v.)\ $Y$ whose expectation is known, then we can construct a new unbiased gradient estimator $\widetilde{\nabla}$ of the function $f$ at point $\x$ as $\widetilde{\nabla} = \nabla_{\mathrm{SGD}} + \alpha(Y-\E Y)$, where $\alpha$ is a constant. If $\alpha = -\frac{\mathrm{Cov}(\nabla_{\mathrm{SGD}},Y)}{\mathrm{Var}[Y]}$, then the variance of the new gradient estimator $\widetilde{\nabla}$ is 
$\mathrm{Var}[\widetilde{\nabla}] =
(1-\kappa^2)
\mathrm{Var}[\nabla_{\mathrm{SGD}}] \le \mathrm{Var}[\nabla_{\mathrm{SGD}}]$, where
$\kappa$
is the Pearson correlation coefficient between $\nabla_{\mathrm{SGD}}$ and $Y$. Usually one may not be lucky enough to have access to such a r.v.\ $Y$ whose covariance with $\nabla_{\mathrm{SGD}}$ is known, therefore the constant $\alpha$ for the control variate needs to be chosen in a sub-optimal or empirical manner.


Finally we want to mention that \cite{SVRG_SGLD_crude} introduce the VR technique to Bayesian inference setting where the objective is the posterior and the gradient estimator is constructed with a minibatch of training data. However, the given convergence guarantee is in terms of mean squared error of statistics evaluated based on posterior, instead of posterior distribution of the parameter. 

\paragraph{Contributions}
This paper discusses the convergence properties of SGLD with variance reduction and shows the ergodicity property of the scheme. Our main contributions upon the prior art is the following:
\begin{itemize}
    \item We provide a better time complexity for the SGLD-VR scheme to converge to a local minimizer than the corresponding result in \cite{SGLD_Gu}. 
    \item We show the ergodicity property of SGLD-VR scheme based on the framework set for SGLD scheme in \cite{Hitting_time_analysis_SGLD_ChenDuTong}.
\end{itemize}

\begin{table*}[t]
\centering
\resizebox{\textwidth}{!}{\begin{tabular}{ lcccc } 
\toprule
Method & w.\ VR & noise magnitude setting & convergence target & Time complexity $m$\\
\midrule
\cite{SGLD_Raginsky} & no & constant & global min.\ & $\dis  \widetilde{\mathcal{O}}\left( \frac{d+1/\delta_0}{\delta_0\eps^4} \right) $  \\
\cite{SGLD_Gu} & yes & constant & global min.$^{\dagger}$\ & $\dis \widetilde{\mathcal{O}}\left( \frac{\sqrt{n}}{\eps^{5/2}} \right)\exp(\widetilde{\mathcal{O}}(d))$\\
\cite{Hitting_time_analysis_SGLG_ZhangLiangCharikar}& no &  constant &  local min.\ & $\dis \mathcal{O}\left( \frac{\Delta_f d^4L^2}{\eps^4} \right)$  \\
This work & yes & diminishing w.\ poly.\ speed & local min.$^*$ & $\dis \mathcal{O}\left(\frac{\Delta_f}{\eps^2}\right) + \exp\left(\mathcal{O}(\eps d)\right) $ \\
\bottomrule
\end{tabular}}
\caption{Comparison between convergence results for variants of LD optimization schemes. $^*$ indicates convergence target is actually a $\eps$-second-order stationary point, which coincides with a local minimizer when $\eps < \sqrt{q}$ under Assumption \ref{as::strict_saddle_ld}. $^\dagger$ means that with the noise magnitude fixed, the optimized empirical error cannot be arbitrarily close to true minimal empirical error, thus returning an \emph{approximate} global minimizer.}
\label{table::ld_convergence_comparison}
\end{table*}

\paragraph{Notation} Bold symbols indicate vectors, for example a vector $\x\in \R^d$, where $d$ stands for the dimension of Euclidean space.  We use $o$ to indicate the starting index of a minibatch,  $\eta_{a:b}$ as the shorthand for $\sum_{i=a}^b \eta_{i}$, and $[n]=\{1,2,\ldots,n\}$.

\section{Algorithm and main results}
The main algorithm is given as Algorithm \ref{alg::LD}. 

\begin{algorithm}[ht]
\caption{Variance reduced stochastic gradient Langevin dynamics (SGLD-VR)
}
\label{alg::LD}
\begin{algorithmic}[1]
\Require initial stepsize $\eta_0>0$ and variance $\rho_0>0$, stepsize decay exponent $\nu \ge 1$, 
batch size $B_b$, epoch length $B_{\e}$
\State Initialize $\x_0=0$, $\xbatch^{(0)} = \x_0$
\State  Define the stepsize and variance sequences:
\begin{equation}
\label{eqn::stepsize_LD}
\eta_t = \frac{\eta_0}{t^\nu} \textrm{ and } \rho_t = \frac{\rho_0}{t^{\nu/2}}.
\end{equation}
\For{$s=0,1,2,\cdots,\frac{T}{B_{\e}}-1$}
\State $\widetilde{\mathbf{w}} = \nabla f(\xbatch^{(s)})$
\For{$l=0,1,\cdots,B_{\e}-1$}
    \State Set index $t=s B_{\e}+l$
    \State Draw $I_t\subset [n]$ of size $|I_t|=B_b$ \Comment{uniformly, with replacement}
    \State Draw $\noise_t\sim \mathcal{N}(0,\mathbf{I})$
    \State  $\gradsvrg_t = \frac{1}{B_b} \sum_{i_t\in I_t} \big( \nabla f_{i_t}(\x_t) - \nabla f_{i_t}(\xbatch^{(s)}) + \widetilde{\mathbf{w}} \big)$ \Comment{gradient estimator}
    \State update $\x_{t+1} = \x_t - \eta_t \gradsvrg_t + \rho_t \noise_t$
\EndFor
\State $\xbatch^{(s)} = \x_{(s+1)B_{\e}}$
\EndFor

\end{algorithmic}
\end{algorithm}


\subsection{Convergence to a first-order stationary point} We start stating the main results with computing the time complexity for the SGLD-VR to converge to an $\eps-$first order stationary point. We define $\x^\star$ to be an $\eps$-first order stationary point (FSP) if $\|\nabla f(\x^\star)\| \le \eps$.

Our first assumption is very standard~\cite{CombettesBook2}:
\begin{assumption}[Lipschitz Gradient]
\label{as::grad_lipschitz_ld}
$f$ is continuously differentiable, and there exists a positive constant $L$ such that for all $\x$ and $\y$, $\|\nabla f(\x) - \nabla f(\y)\|\le L\|\x-\y\|$.
\end{assumption}

\begin{theorem}
\label{thm::first_order_convergence}
 Under Assumption \ref{as::grad_lipschitz_ld}, for any $p\in(0,1)$, then with probability at least $1-p$, the time complexity for the LD described in Algorithm \ref{alg::LD} to converge to an $\eps$-first order stationary point $\x^\star$ is $\displaystyle \mathcal{O}\left( \frac{\Delta_f d}{ \eps^2 p} \right)$, where $\Delta_f = f(\x_0) - f(\x^\star)$.
\end{theorem}

\subsection{Ergodicity} In this work we show that the discretized variance reduced LD (SGLD-VR) has an ergodic property which gives the iteration process the potential of exploring wider space, thus with positive possibility of traversing through the global optimal point. We make the following regularization assumptions.

The following assumption says essentially that $f$ is bounded below by a known value (and without loss of generality, we can assume $f$ is non-negative). This automatically holds for ERM problems when the loss function is non-negative, as is typical.
\begin{assumption}[Nonnegative objective]
\label{assp::nonnegativity}
The objective function is nonnegative.
\end{assumption}

The next assumption is more complex and we discuss it in Remark~\ref{remark:assumption}.
\begin{assumption}[Regularization conditions] 
\label{assumption::regularization_conditions}
There exist nonnegative constants $\mu_1$, $\mu_2$ and $\psi_1$, $\psi_2$ such that for all $\x\in\R^d$, 
\begin{align}
\label{ineqn::ergo_regularisation_1}
    \|\nabla f(\x)\|^2 &\ge \mu_1 f(\x) - \psi_1\\
\label{ineqn::ergo_regularisation_2}
    \|\x\|^2 &\le \mu_2 f(\x) + \psi_2
\end{align}
\end{assumption}

\begin{remark}\label{remark:assumption}
We make the same regularization assumptions as in \cite{Hitting_time_analysis_SGLD_ChenDuTong}. A similar regularization condition to \eqref{ineqn::ergo_regularisation_1} commonly used in previous literature is the \textit{$(m,b)$-dissipative condition} \cite{Mattingly_ergodicity_2002,SGLD_Raginsky,SGLD_Gu,Hitting_time_analysis_SGLG_ZhangLiangCharikar}, which reads that there exist positive constants $m$ and $b$ such that for all $\x\in\R^d$, $\<\nabla f(\x),\x\>\ge m\|\x\|^2-b$. \cite{dong2020replica} show that the dissipative condition implies \eqref{ineqn::ergo_regularisation_1}, which renders the assumption \eqref{ineqn::ergo_regularisation_1} weaker. 
Another interpretation of \eqref{ineqn::ergo_regularisation_1}, in conjunction with Assumption~\ref{assp::nonnegativity}, is that this is a slightly weaker version of the Polyak-{\L}ojasiewicz (PL) inequality~\cite{Karimi2016LinearCO}; choosing $\psi_2$ to be the minimal value of $f$ gives the PL inequality, but the PL inequality itself is stronger since it implies that any stationary point (i.e., where $\nabla f(x)=0$) is globally optimal. 
Equation \eqref{ineqn::ergo_regularisation_2} implies that $f$ is supercoercive~\cite{CombettesBook2}, and in particular coercive, and thus has bounded level sets.

Consider the function $f(\x) = \sigma(\mathbf{A}\x) + \gamma \|\x\|_2^2 + C$, which describes the connection between one neuron and the layer below it in a feedforward neural network with coefficient matrix $\mathbf{A}$, the activation function $\sigma$ as $\tanh$ or sigmoid, and a  Tikhonov regularization term with magnitude $\gamma$ and a constant $C$. This is an example which satisfies Assumptions \ref{as::grad_lipschitz_ld}, \ref{assp::nonnegativity}, and \ref{assumption::regularization_conditions}. Examples with more types of activation functions such as ReLu and more types of regularization terms such as $\ell_1$, or extensions to multilayer feedforward networks or convolutional neural networks can also be constructed if they are defined region-wise to cater for near-origin behavior and far-field behavior in the regularization assumption \ref{assumption::regularization_conditions} respectively.
\end{remark}

In Theorem~\ref{thm::ergodicity} we show that there is a nonzero probability that the LD iteration will eventually visit any fixed point within a level set of interest.

\begin{theorem}[Ergodicity]
\label{thm::ergodicity}
Under assumptions \ref{as::grad_lipschitz_ld}, \ref{assp::nonnegativity}, and \ref{assumption::regularization_conditions},
with the same parameter setting as in Lemma \ref{lemma::recurrence}, for any accuracy  $\widetilde{\eps}>0$, failure probability $p>0$,  and any point $\mathbf{s}\in\R^d$ which locates in the level set $\{\x: f(\x)\le\mathcal{O}(\widetilde{\eps})\}$, there is a finite time horizon 
\begin{equation}
T = \widetilde{\mathcal{O}}\left(\frac{1 + \ln f(\x_0)   + \frac{(d\|\mathbf{s}\| + \widetilde{\varepsilon}) ^d }{ \widetilde{\eps}\big( (\frac{4}{\sqrt{2\pi}}-1)\mathrm{e}^{-1/2} \widetilde{\varepsilon} \big)^d}}{\widetilde{\eps} p\mu_1\big(\psi_1+ 2\eta_0L^3 \frac{\mu_2 f(\x_0) + 2\psi_2}{B_\e} + \frac{\rho_0^2}{\eta_0} Ld\big)} \right)
\end{equation} such that
\begin{equation}
\label{ineqn::ergodicity_main_result}
    \pr(\|\x_t - \mathbf{s}\|\le \widetilde{\eps} \,\textrm{ for some }\, t<T) \ge 1-p
\end{equation}
\end{theorem}

\subsection{Convergence to an $\eps$-second-order stationary point}
An $\eps$-second-order stationary point is a more restrictive type of $\eps$-first order stationary point, and is more likely to be an actual local minimizer. 
\begin{definition} 
\label{def::essp}
Consider a smooth function $f(\x)$ with continuous second order derivative. 
A point $\x$ is an $\eps$-second-order stationary point if 
\begin{equation}
    \|\nabla f(\x)\| \le \eps ~\textrm{ and }~ \lambda\big(\nabla^2 f(\x)\big)_{\min} \ge -\eps^2
\end{equation}
where $\lambda(\cdot)_{\min}$ is the smallest eigenvalue.
\end{definition}
We make the strict saddle assumption which is common in nonconvex optimization literature~ \cite{Ge_EscapeSaddle,lee2016gradient,jin2017escape,constrainedEscapeSaddle18,Lee2019_FirstOrderAvoidSaddle,AvoidSaddle_zeroorderMethod19,HeavyBall_escapesaddle19,liu2018envelope,SSRGD_Li19,Huang_PPD}; i.e.,
\begin{assumption}[Strict saddle]
\label{as::strict_saddle_ld}
There exists a constant $q>0 $ such that for all first-order stationary points $\fsp$,  we have
\[
|\lambda(\nabla^2 f(\fsp))| \ge q >0.
\]
\end{assumption}

\begin{assumption}[Lipschitz Hessian]
\label{as::hessian_lipschitz_ld}
$f$ is twice continuously differentiable, and there exists a positive constant $L_2$ such that for all $\x$ and $\y$, $\|\nabla^2 f(\x) - \nabla^2 f(\y)\|\le L_2\|\x-\y\|$.
\end{assumption}

\begin{theorem}
\label{thm::second_order_convergence}
Under Assumptions \ref{as::grad_lipschitz_ld}, \ref{as::strict_saddle_ld} and \ref{as::hessian_lipschitz_ld}, setting the stepsize decay parameter $\nu\in[1,2]$ and $\rho_0=\mathcal{O}(\eps)$, with probability $\mathcal{O}\left( \frac{\eps^{d-1}}{ \Gamma(\frac{d-2}{2}) L^{d-1}q^{d-1} } \right)$, the time complexity for the LD described in Algorithm \ref{alg::LD} to converge to an $\eps$-second order stationary point $\x^\star$ is $\mathcal{O}\left(\frac{\Delta_f}{\eps^2}\right) + \exp\left(\mathcal{O}(\eps d)\right)$, where $\Delta_f = f(\x_0) - f(\x^\star)$.
\end{theorem}

\begin{table*}[t]
\centering
\resizebox{\textwidth}{!}{\begin{tabular}{ lccccc } 
\toprule
Method & Bounded & Grad. Lip. & Hess. Lip. & Regularization & Other assumptions\\
\midrule
\cite{SGLD_Raginsky} & $f$ and $\|\nabla f\|$ & yes & no & $(m,b)$-dissipative &  \shortstack{ 1) stoch.\ grad.\ sub-exp.\ tails\ \\ 2) init.\ pt.\  sub-Gauss.\ tails } \\
\cite{SGLD_Gu} & none & yes & no & $(m,b)$-dissipative & none \\
\cite{Hitting_time_analysis_SGLG_ZhangLiangCharikar}& $\|\nabla f\|$ and $\|\nabla^2 f\|$ &  yes &  yes & $(1,0)$-dissipative & grad.\ sub-exp.\ tails  \\
This work & none & yes & yes & Assumption \ref{assumption::regularization_conditions} & strict saddle  \\
\bottomrule
\end{tabular}}
\caption{Comparison between assumptions made for variants of LD optimization schemes. The Hessian Lipschitz assumption is used only for claims about second-order convergence.}
\label{table::assumptions_summary}
\end{table*}

\section{Proof Sketch}

\subsection{Convergence to a first-order stationary point}
We first bound the expectation of the square of the gradient norm in a minibatch step of SGLD-VR. 
To estimate the time needed to converge to a first-order stationary point (FSP), we compute the dependence of the gradient norm bound on the  iteration count $t$. The quantity that plays a central role in the argument is the Lyapunov function, which is essential in constructing the upper bound for gradient norm and connects the argument between successive minibatches.

\begin{lemma}[Bound of variance of SVRG gradient estimator \cite{Reddi_SVRG_first_order_2016}] In an epoch, 
the SVRG gradient estimator satisfies
\begin{equation}
\label{ineqn::variance_gradsvrg_bound}
\E[\|\gradsvrg_t\|^2] \le 2\E[\|\nabla f(\x_t)\|^2] + 2\frac{L^2}{B_\e}\E[\|\x_t-\xbatch\|^2].
\end{equation}
\label{lemma::variance_gradsvrg}
\end{lemma}

Adapting the framework in \cite{Reddi_SVRG_first_order_2016} for the LD setting, the following lemma bounds the expectation of the gradient norm for the SGLD-VR iteration sequence in a minibatch:
\begin{lemma}
\label{lemma::lyapunov_bound_gradient}
Define the weight sequence $(c_t)$ recursively as $c_t = c_{t+1}(1+\beta_t\eta_t + 2\frac{\eta_t^2 L^2}{B_\e}) + \frac{\eta_t^2 L^3}{B_e}$ with $c_{B_\e} = 0$, and then define the Lyapunov function $R_t = \E[f(\x_t)+c_t\|\x_t-\xbatch\|^2]$ for each epoch. Define the normalization sequence $\gamma_t = \eta_t - \frac{c_{t+1}}{\beta_t}\eta_t - \eta_t^2 L - 2c_{t+1}\eta_t^2$ with $\eta_t$ and $\beta_t>0$ set to ensure $\gamma_t > 0$.  Under Assumption \ref{as::grad_lipschitz_ld}, inside an epoch,
\[
\E[\|\nabla f(\x_t)\|^2] \le \frac{R_t-R_{t+1}}{\gamma_t} + \left(\frac{L}{2}+c_{t+1}\right)\frac{d\rho_t^2}{\gamma_t}.
\]
\end{lemma}
Remark~\ref{remark:sequence} in the supplementary material shows that there always exists choices of $\eta_0$ and $\nu$ (hence $\eta_t$ via \eqref{eqn::stepsize_LD}) and $(\beta_t)$ to ensure $\gamma_t> 0$.

Now we use the bound of the gradient norm within a minibatch to build the norm bound of the SGLD-VR gradient estimator for the whole iteration in the following lemma, with which one can derive the time complexity for the SGLD-VR scheme to converge to a FSP as in Theorem \ref{thm::first_order_convergence}: 
\begin{lemma}\label{lemma::grad_norm_bound_entire_iter}
Let $\Bar{\gamma} = \min_{0\le t\le T-1} \gamma_t$ where $\gamma_t$ is defined in the previous lemma, and $\nu>0$. Then under Assumption \ref{as::grad_lipschitz_ld},
\begin{equation}
\label{ineqn::grad_bound}
\E[\|\nabla f(\x_a)\|^2] \le \frac{f(\x_0)-f(\x^\star)}{T\Bar{\gamma}} + \frac{d}{\Bar{\gamma}} \left(\frac{L}{2}+c_0\right)\frac{C_0}{T^{\nu}},
\end{equation}
where $\x_a$ is randomly chosen from the entire iterate sequence and $C_0$ is a universal constant.
\end{lemma}


\subsection{Ergodicity}
The ergodicity argument is comprised of two parts: recurrence and reachability. 
\paragraph{Recurrence}
The LD term in the optimization scheme, due to its random-walk nature, is the key for the  reachability argument. In this section we follow the framework of \cite{Hitting_time_analysis_SGLD_ChenDuTong} while giving new specific proofs.

We first show that with Langevin dynamics, the iteration process will visit sublevel sets of interest, for instance the collection of compact neighborhoods of all local minimums, infinitely many times. Lemma \ref{lemma::recurrence} is the first pillar to establish the ergodicity result. 
In its proof, we first give a more explicit characterization of function value decrease between two successive SGLD-VR updates than the characterization using the Lyapunov function $R_t$ for the discussion of convergence to first-order stationary points in lemma \ref{lemma::lyapunov_bound_gradient}. Next, we construct a supermartingale involving the objective function value and iteration count. Through the introduction of a stopping time sequence which records the time of the iteration visiting targeted sublevel sets, one can establish the expectation of any entry in this stopping time sequence, thus proving the lemma.

Our main lemma is 
Lemma~\ref{lemma::recurrence} 
where we 
give an explicit upper bound of the expected time of visiting a given level set for the $j$-th time ($j\ge 1$).

\begin{lemma}[Recurrence]
\label{lemma::recurrence}
For a fixed $\delta>0$, let $n_0$ be the index such that $\eta_{n_0}\le \delta$, and $n_k$ be the sequence of iteration index $n_{k+1} = \min_s\{ s : s>n_k, \,  \eta_{n_k:s} \ge \delta   \}$. 

Under Assumptions \ref{as::grad_lipschitz_ld}, \ref{assp::nonnegativity} and \ref{assumption::regularization_conditions}, there exists a constant $C_1$ such that for constants $\alpha = 1-2\exp(-(1-C_1)\mu_1\delta)$, $B = 2( \psi_1 + \frac{2\eta_{n_0} L^3}{B_\e}\big( \mu_2 f(\x_0) + 2\psi_2 \big) + \frac{\rho_0^2 Ld}{2\eta_0})$, $K = \frac{\ln\frac{f(\x_{n_0})}{\delta B}}{(1-C_1)\mu_1\delta}$, the stopping time sequence $\{\tau_k\}$ defined as $\tau_{0}=K$ and $\tau_{k+1} = \min\{t: t\ge \tau_k + 1, f(\x_{n_t})\le 2\delta B\}$ satisfies 
\begin{equation}
    \E[\tau_j] \le \frac{4}{\alpha} + K + j\left(\frac{1}{2\alpha\delta }+1\right).
\end{equation}
\end{lemma}

\begin{remark} 
As we have assumed that $f\ge 0$ which 
is common for ERM problems since the loss function is usually non-negative, 
it is desirable that $f(\x_{n_t})$ goes to 0 as the iteration proceeds. Thus, the choice of $\delta$ for analytical purposes would be $\dis \delta \propto \frac{\widehat{\varepsilon}}{B} $
for some $\widehat{\varepsilon}$-target level one deems appropriate.
\end{remark}

\paragraph{Reachability}
As lemma \ref{lemma::recurrence} shows, when the expected time for the iterates to revisit a certain level set of interest for $j$-th time is finite ($j$ is any positive integer), we call such a level set recurrent. We show that when the SGLD iteration sequence starts from a recurrent compact set, there is a positive possibility for the sequence to visit every nearby first-order stationary point.

Lemma \ref{lemma::brownian_ergodicity} is the core lemma to establish the ergodicity result for the LD optimization scheme. The core idea behind its proof is to leverage the exploratory potential of a \textit{radial} Brownian motion process to show that there is a non-trivial probability for the Gaussian noise accumulation in the LD scheme to visit a pre-designated point in space.

\begin{lemma}[Ergodicity due to Brownian motion]
\label{lemma::brownian_ergodicity}
Given any sequence $a_k > 0$,  
let $\z_k = \sum_{i=1}^k \rho_0\sqrt{a_i}\noise_i$ where $\noise_i \sim\mathcal{N}(0,I_d)$, for any target vector $\z^\star$ and distance $r$, there exists a positive function $p_1$ such that  
\begin{multline}
\pr(\|\z_n-\z^\star\|\le r, \, \|\z_k\| \le  \|\z^\star\|+r \,\, \forall k=1,\cdots, n )\\ \ge p_1(r,\rho_0,t_n,\z^\star)
\end{multline}
with $p_1(0,\rho_0,t_n,\z^\star) = p_1(r,0,t_n,\z^\star) = 0$. 
\end{lemma}

We leverage Lemma \ref{lemma::brownian_ergodicity} to show the reachability of SGLD-VR scheme, which is the second pillar to establish the ergodicity result. The core idea behind the proof of Lemma \ref{lemma::reachability} is to balance the influence on the iterates from gradient descent and Gaussian noise accumulation respectively, and show that the exploratory potential behind the Gaussian noise accumulation will fulfill the desired property of reachability.
\begin{lemma}[Reachability]
\label{lemma::reachability}
Assume the same stepsize batch setting as in Lemma \ref{lemma::recurrence} and the gradient Lipschitz condition in Assumption \ref{as::grad_lipschitz_ld}, with respect to an arbitrary target point $\mathbf{s}$ in the level set $\{\x: f(\x)\le \eps\}$.  there is a constant $C_{21} \propto dL$ and constant $C_{22}$ such that for any $\eps>0$,
\begin{equation}
    \pr\left( \| \x_{n_{i+1}} - \mathbf{s} \| \le \widetilde{\eps} \right) > \frac{1}{2}p_1(\widetilde{\eps},\rho_0,t_{n_{i+1}},\mathbf{s})
\end{equation}
where $\widetilde{\eps} = \eps + \delta C_{21}+ 2\delta \sqrt{C_{22}}$.
\end{lemma}
Theorem \ref{thm::ergodicity} follows by bounding the probability $\pr(\tau_\star>T)$ for some predesignated $T$, where we define $\tau_* = \min\{t: t>0, \|\x_{n_t} - \mathbf{s}\| \le \widetilde{\eps}\}$. With the marker sequence of iteration index $n_{k} = \min_s\{ s : s>n_{k-1}, \,  \eta_{n_{k-1}:s} \ge \frac{\widetilde{\eps}}{2B} \}$ and the auxiliary stopping time sequence $\tau_0 = K$, $\tau_{t+1} = \min\{t: t\ge \tau_k + 1, f(\x_{n_t}) \le \widetilde{\eps}\}$, where $B$ and $K$ are objective-specific constants, one writes $\pr(\tau_* \ge T) = \pr(\tau_* \ge T, \tau_J >T) + \pr(\tau_* \ge T, \tau_J <T)$ and bound each term by Lemma \ref{lemma::recurrence} and Lemma \ref{lemma::reachability} respectively.


\subsection{Convergence to a second-order stationary point}
By far in the literature there are two common ways to argue the convergence to second-order stationary points (SSP)
\begin{itemize}
    \item show that $f(\x_T) - f(\x_0) < \Delta_f$ with probabilistic guarantee to ensure the continual function value decrease at saddle point (\cite{jin2017escape})
    \item show that $\|\x_T - \x^\star\|$ decreases in the probabilistic sense as $T$ increases (\cite{Kleinberg_SGD_escape_local_minima}). 
    
    The time complexity of this approach has the exponential dependency on the inverse of the error tolerance. So in this work we resort to the previous approach.
\end{itemize}

The argument to show sufficient function value decrease from a FSP in \cite{jin2017escape} uses two iterate sequences to demonstrate the continual function value decrease at saddle point. Now that the noise is injected at every iteration, the geometric intuition that the trapping region is thin plus the probabilistic argument should be able to give a similar proof.

In the LD setting we exploit the property of Brownian motion to show the escape from saddle point, i.e. to characterize the perturbed iterate has high probability in the direction of descent, 
\[
(\x_{t} - \x_{\textrm{fsp}})^\trans \nabla^2 f (\x_\textrm{fsp}) (\x_{t} - \x_{\textrm{fsp}}) \le -\zeta 
\]

The proof contains four steps:\begin{enumerate}
    \item We show that $\Delta_i  := \sum_{l=n_{i}}^{n_{i+1}-1} \sqrt{\eta_i} \noise_i$ will lead to saddle point escape, i.e. $\Delta_i^\trans \nabla^2 f(\x_{\textrm{fsp}}) \Delta_i \le -\zeta $. Specifically, show that $\Delta_i$ has projection on the direction of $\lambda_{\mathrm{min}}$ more than $\zeta $ with high probability, which exploits the property of Brownian motion and the idea that the trapping region is thin when faced with LD (\cite{jin2017escape}).
    \item We show that when $\x \in \mathcal{U}(\x_{\mathrm{fsp}},r)$ where $\|\Delta_j\|<r$ for $j=n_i, n_i+1,\cdots, n_{i+1}-1$, $\|\nabla f(\x)\|<\eps$, thus the first order expansion does not contribute to function value change.
    \item We show that the update $\x' = \x + \Delta_i$ will lead to function value decrease, thus the SGLD algorithm has to terminate, thus converging to SSP.
    \item Compute $\tau_{\mathrm{SSP}}$ by taking account of the time needed for escaping saddle points and the time needed for achieving sufficient function value decrease.
\end{enumerate}

\section{Conclusion}
In this paper we consider the application of the scheme stochastic gradient Langevin dynamics with variance reduction on minimizing nonconvex objectives, prove the probabilistic convergence guarantee to local minimizers, and prove corresponding ergodicity property of the scheme which leads to non-trivial probability for the scheme to visit global minimizers.

\section*{Acknowledgments}
This material is based upon work supported by the National Science Foundation under grant no.\ 1819251.

Zhishen Huang thanks Manuel Lladser for helpful discussions on properties of Brownian motion. 

The research presented in this paper was performed when ZH was affiliated with Department of Applied Mathematics, University of Colorado Boulder.

\bibliographystyle{apalike} 
\bibliography{IEEEabrv,references,newReferences}

\onecolumn
\appendix 
\section*{Supplementary Material}
\input{first_order_convergence.tex}
\input{ergodicity.tex}
\input{second_order_convergence.tex}

\end{document}

%% file: first_order_convergence.tex
\subsection{Proofs of first-order stationary point convergence property}
In this section we  prove  the result for first-order convergence property (theorem \ref{thm::first_order_convergence}) as well as the needed lemmas.


\begin{lemma}
[Repeat of lemma \ref{lemma::lyapunov_bound_gradient}]
Define the weight sequence $\{c_t\}$ recursively as $c_t = c_{t+1}(1+\beta_t\eta_t + 2\frac{\eta_t^2 L^2}{B_\e}) + \frac{\eta_t^2 L^3}{B_e}$ with $c_{B_\e} = 0$, and then define the Lyapunov function $R_t = \E[f(\x_t)+c_t\|\x_t-\xbatch\|^2]$ for each epoch. Define the normalization sequence $\gamma_t = \eta_t - \frac{c_{t+1}}{\beta_t}\eta_t - \eta_t^2 L - 2c_{t+1}\eta_t^2$ with $\eta_t$ and $\beta_t>0$ set to ensure $\gamma_t > 0$.  Under Assumption \ref{as::grad_lipschitz_ld}, inside an epoch,
\[
\E[\|\nabla f(\x_t)\|^2] \le \frac{R_t-R_{t+1}}{\gamma_t} + \left(\frac{L}{2}+c_{t+1}\right)\frac{d\rho_t^2}{\gamma_t}.
\]
\end{lemma}

\begin{proof}
We find upper bounds to the Lyapunov functions $R_t$ in terms of the negative norm of the SVRG gradient estimator, thus proving the lemma. We bound the two terms in the Lyapunov functions respectively. For notational simplicity let 
$\nabla f(\x_t) = \nabla_t = \E_{I_t}[\gradsvrg_t]$.

For the first term $f(\x_{t+1})$ in the Lyapunov function,
using Prop.~\ref{prop:descent},
\begin{align*}
\E[f(\x_{t+1})] 
&= \E\bigg[ f(\x_t)-\eta_t\|\nabla_t\|^2 + \frac{L}{2}\big(\eta_t^2\|\gradsvrg_t\|^2 + \rho_t^2\|\noise_t\|^2\big) \bigg].
\end{align*}

For the second term $\|\x_{t+1} -\xbatch\|$, as $\<\nabla_t,\xbatch-\x_t\> \stackrel{\text{CS}}{\le} \|\nabla_t\|\|\x_t-\xbatch\| \stackrel{\text{Young}}{\le} \frac{1}{2\beta_t}\|\nabla_t\|^2 + \frac{\beta_t}{2}\|\x_t-\xbatch\|^2$,
\begin{align*}
    \E[\|\x_{t+1} - \xbatch\|^2] &= \E[\|\x_{t+1} - \x_t + \x_t -  \xbatch\|^2] = \E[ \|\x_{t+1}-\x_t\|^2 + \|\x_t-\xbatch\|^2 + 2\<\x_{t+1}-\x_t, \x_t-\xbatch \>]\\
    &= \E\big[ \eta_t^2 \|\gradsvrg_t\|^2 + \rho_t^2\|\noise_t\|^2 + \|\x_t-\xbatch\|^2 + 2 \eta_t \<\nabla_t , \xbatch-\x_t\> \big]\\
    &\le \E \big[ \eta_t^2  \|\gradsvrg_t\|^2 + \rho_t^2 \|\noise_t\|^2 + (1+\eta_t\beta_t)\|\x_t-\xbatch\|^2 +\frac{\eta_t}{\beta_t} \|\nabla_t\|^2 \big]
\end{align*}

Putting these two terms together into $R_{t+1}$, we have
\begin{align*}
    R_{t+1} &\le \E\left[
    f(\x_t) + \left(\frac{\eta_t c_{t+1}}{\beta_t}-\eta_t\right)\|\nabla_t\|^2 + \left(\frac{L}{2}+c_{t+1}\right) ( \eta_t^2 \|\gradsvrg_t\|^2 + \rho_t^2 \|\noise_t\|^2) + (1+\eta_t\beta_t)c_{t+1}\|\x_t-\xbatch\|^2 \right] \\
    &\stackrel{\eqref{ineqn::variance_gradsvrg_bound}}{\le}
    \E\Big[f(\x_t) 
    + \left(\frac{\eta_t c_{t+1}}{\beta_t}-\eta_t + (L+2c_{t+1})\eta_t^2\right)\|\nabla_t\|^2 
    + \left(\frac{L}{2}+c_{t+1}\right)  \rho_t^2 \|\noise_t\|^2\\
    &\quad \quad\quad + \big((1+\eta_t\beta_t)c_{t+1} + (L+2c_{t+1})\frac{\eta_t^2 L^2}{B_\e} \big)\|\x_t-\xbatch\|^2 \Big]\\
    &= \E\left[f(\x_t) - \gamma_t\|\nabla_t\|^2 + \left(\frac{L}{2}+c_{t+1}\right)\rho_t^2\|\noise_t\|^2 + c_t\|\x_t-\xbatch\|^2 \right]\\
    &= R_t- \E\big[\gamma_t\|\nabla_t\|^2] + \E\left[\left(\frac{L}{2}+c_{t+1}\right)\rho_t^2\|\noise_t\|^2\right] \\
    &= R_t- \E\big[\gamma_t\|\nabla_t\|^2] + \left(\frac{L}{2}+c_{t+1}\right)\rho_t^2d.
\end{align*}
We set $\{\beta_t\}$ and $\eta_0$ properly (see the remark below this proof) such that $-\gamma_t = \frac{\eta_t c_{t+1}}{\beta_t}-\eta_t + (L+2c_{t+1})\eta_t^2 \le 0$ for all $t=0,1,\cdots,B_\e-1$. 
This can always be achieved as $c_t$ is a decreasing sequence and $c_t$ is negatively related to $\beta_t$. Then 
\begin{equation}
\E[\gamma_t\|\nabla_t\|^2] \le - R_{t+1} + R_t + \left(\frac{L}{2}+c_{t+1}\right)\rho_t^2d.
\end{equation}
\end{proof}

\begin{remark} \label{remark:sequence}
We show that the $\eta_0$ and $\{\beta_t\}$ sequence setting in the end of the proof of lemma \ref{lemma::lyapunov_bound_gradient} always exists. For now we can assume $\beta_t = \widetilde{\beta}$ is a constant. Then we can define an upper bound sequence for $\{c_t\}$ as
\[
\widetilde{c}_t = \widetilde{c}_{t+1}(1+\widetilde{\beta}\eta_0+2\frac{\eta_0^2 L^2}{B_\e}) + \frac{\eta_0^2 L^3}{B_\e} 
\]
with $\widetilde{c}_{B_\e} = 0$. Then, $c_t\le \widetilde{c}_{t}$ for $1 \le t\le B_\e$. Consequently, for expression simplicity assuming $q = 1+\widetilde{\beta}\eta_0+2\frac{\eta_0^2 L^2}{B_\e}$ and $D = \frac{\frac{\eta_0^2 L^3}{B_\e}}{\widetilde{\beta}\eta_0+\frac{2\eta_0^2 L^2}{B_\e}} = \frac{\frac{\eta_0 L^3}{B_\e}}{\widetilde{\beta}+\frac{2\eta_0 L^2}{B_\e}}$, we have 
\[
\frac{\widetilde{c}_t+D}{\widetilde{c}_{t+1}+D} = q. 
\]
It follows that $\frac{1}{q^{B_\e}}(\widetilde{c}_0+D) = \widetilde{c}_{B_\e}+D = D$, and $\widetilde{c}_0 = (q^{B_\e}-1) D$. We need to set $\widetilde{\beta}$ in a way such that $\gamma_t >0$ for all $1\le t\le B_\e$. As
\[
\gamma_t \ge \left(1 -\frac{\widetilde{c}_0}{\widetilde{\beta}} - \eta_0 L - 2 \widetilde{c}_0 \eta_0 \right)\eta_t \stackrel{\texttt{need}}{>} 0 ,
\]
a sufficient condition to assure the second inequality above is
\begin{equation}
\label{ineqn::c0_eta_0_cond}
\widetilde{c}_0\left(\frac{1}{\widetilde{\beta}} + 2\eta_0 \right) + \eta_0 L < 1
\end{equation}
Let $\widetilde{\beta}\eta_0$ be small while $\widetilde{\beta}>1$, then the l.h.s.\ of \eqref{ineqn::c0_eta_0_cond} is of the order $B_\e \eta_0 \frac{\frac{\widetilde{\beta}\eta_0 L^3}{B_\e}}{\widetilde{\beta}^2 + 2 \frac{\widetilde{\beta}\eta_0 L^2}{B_\e}}$, which can ensure \eqref{ineqn::c0_eta_0_cond} to hold.
\end{remark}

\begin{lemma}[Repeat of lemma \ref{lemma::grad_norm_bound_entire_iter}]
Let $\Bar{\gamma} = \min_{0\le t\le T-1} \gamma_t$ where $\gamma_t$ is defined in the previous lemma, and $\nu>0$. Then under Assumption \ref{as::grad_lipschitz_ld},
\begin{equation}
\E[\|\nabla f(\x_a)\|^2] \le \frac{f(\x_0)-f(\x^\star)}{T\Bar{\gamma}} + \frac{d}{\Bar{\gamma}} \left(\frac{L}{2}+c_0\right)\frac{C_0}{T^{\nu}},
\end{equation}
where $\x_a$ is randomly chosen from the entire iterate sequence and $C_0$ is a universal constant.
\end{lemma}

\begin{proof}
We set $c_{B_\e} = 0$ so that $R_0^{(\alpha)} = f(\x_0^{(\alpha)})$ and $R_{B_\e}^{(\alpha)} = f(\x_{B_\e}^{(\alpha)})$ for the fixed epoch $\alpha$. Per line 10 in Algorithm \ref{alg::LD}, the ending point of the previous epoch is the starting point of the next epoch, i.e., $\x_0^{(\alpha)} = \x_{B_\e}^{(\alpha-1)}$. Summing up all the iteration steps in each epoch, we have
\[
\sum_{\alpha = 0}^{\frac{T}{B_\e}-1} \sum_{l=0}^{B_\e-1}  \E[\nabla f(x_l^{(\alpha)})] \le \frac{ f(\x_0)-f(\x_T) }{\bar{\gamma}} + \frac{\E[\|\noise\|^2]}{\bar{\gamma}} \sum_{t=0}^{T-1}\left(\frac{L}{2}+c_{(t \textrm{ mod } B_\e) +1}\right)\rho_t^2 
\]

When $\rho_t$ is set as $\mathcal{O}(\frac{1}{t^{\nu/2}})$ where $\nu\ge 1$, as $c_t$ is bounded w.r.t.\ a fixed epoch, $\sum_{t=0}^{T-1} \rho_t^2 = \mathcal{O}(T^{1-\nu})$. (The $\nu = 1$ case leads to logarithmic growth of summation of $\rho_t^2$, which does not affect the following result.) Then consider the LHS of the inequality as the average over all iterates, then 
\begin{align}
\label{ineqn::aver_gradient_bound}
\E[\|\nabla f(\x_a)\|^2] &\le \frac{f(\x_0)-f(\x^\star)}{T\Bar{\gamma}} + \frac{\E[\|\noise\|^2]}{T\Bar{\gamma}} \sum_{t=0}^{T-1}\left(\frac{L}{2}+c_{(t \textrm{ mod } B_\e) +1}\right)\rho_t^2 \nonumber\\
&\le \frac{f(\x_0)-f(\x^\star)}{T\Bar{\gamma}} + \frac{d}{T\Bar{\gamma}}\left(\frac{L}{2}+c_0\right)\sum_{t=0}^{T-1}(\frac{L}{2}+c_0)\rho_t^2 \nonumber\\
&= \frac{f(\x_0)-f(\x^\star)}{T\Bar{\gamma}} + \frac{d}{\Bar{\gamma}} \left(\frac{L}{2}+c_0\right)\frac{C_0}{T^{\nu}}.
\end{align}

\end{proof}

\begin{proof}[Proof of Thm.~\ref{thm::first_order_convergence}]
Per \eqref{ineqn::aver_gradient_bound}, we see that the time complexity for the LD to converge to an $\eps$-first order stationary point is $\mathcal{O}\big( \frac{\Delta_f d}{\bar{\gamma} \eps^2} \big)$. Another way to phrase the time complexity is through the hitting time of LD to a first-order stationary point (fsp) $\tau_{\textrm{fsp}}$. To estimate the expected time for the iteration sequence to enter a fsp neighborhood,
\begin{align*}
\pr(\tau_{\textrm{fsp}}>T) 
= \pr(\|\nabla f(\x_t)\|> \eps, \,\, \forall t\le T) &\le  \pr\bigg(\frac{1}{T} \sum_{t=1}^T\|\nabla f(\x_t)\|> \eps \bigg) \\
&\le \frac{\E[\frac{1}{T} \sum_{t=1}^T\|\nabla f(\x_t)\|]}{\eps} \\
&= \frac{\E[\|\nabla f(\x_a)\|]}{\eps} \le  \frac{\sqrt{\E[\|\nabla f(\x_a)\|^2]}}{\eps},
\end{align*}
where the 2nd inequality is due to Markov's inequality, and the expectation in the final line is taken over choosing $a$ uniformly from $\{1,\ldots,T\}$ in addition to the other random variables, and the final inequality is Jensen's inequality.

Thus, using Lemma~\ref{lemma::grad_norm_bound_entire_iter},
\begin{equation}
\pr(\tau_{\textrm{fsp}}>T) 
\le \frac{  1}{\eps}\sqrt{
\frac{\Delta_f}{T\Bar{\gamma}} + \frac{d}{\Bar{\gamma}} \left(\frac{L}{2}+c_0\right)\frac{C_0}{T^{\nu}}
} \stackrel{\textrm{let}}{ = } p,
\end{equation}
where $p$ is the failure probability.
As $\bar{\gamma}$ is a positive constant independent of $d,\eps$ and $T$, the equation above transforms into $ \frac{\Delta_f}{T\Bar{\gamma}} + \frac{d}{\Bar{\gamma}} \left(\frac{L}{2}+c_0\right)\frac{C_0}{T^{\nu}} = \eps^2 p$. As $\nu\ge 1$, $T = \mathcal{O}\left(\frac{\Delta_f d}{\bar{\gamma}\eps^2 p}\right)$.
\end{proof}

%% file: ergodicity.tex
\subsection{Proofs of ergodicity properties}
Let $\mathcal{F}_{t}$ be the filtration generated by $(\x_0,\ldots,\x_{t})$ and associated random variables ($I_t$ and $\noise_t$) for $(\x_{t})$ the sequence from Algo.~\ref{alg::LD}.
We write $\E[ \cdot \mid \mathcal{F}_{t} ]$ as just $\E[ \cdot ]$ when the conditioning is clear from context.

We start with a simple proposition that will be used in several of the proofs.
\begin{proposition}[variant of the Descent Lemma] \label{prop:descent}
For $\x_t$ generated via Algo.~\ref{alg::LD}, under the Lipschitz assumption \ref{as::grad_lipschitz_ld}, 
then the expectation conditioned on $\mathcal{F}_{t}$
satisfies 
\begin{align} \label{eq:descent}
\E[f(\x_{t+1})] &\le \E\bigg[f(\x_t) + \<\nabla f(\x_t), \x_{t+1}-\x_t \> + \frac{L}{2}\|\x_{t+1}-\x_t\|_2^2\bigg]\notag\\
&= \E\bigg[ f(\x_t)-\eta_t\|\nabla_t\|^2 + \frac{L}{2}\big(\eta_t^2\|\gradsvrg_t\|^2 + \rho_t^2\|\noise_t\|^2\big) \bigg]
\end{align}
\end{proposition}
\begin{proof}
The first inequality uses the $L$-smoothness of function $f$ and the second equality uses the SVRG update in algorithm \ref{alg::LD} ($\x_{t+1}-\x_t = -\eta_t\gradsvrg_t + \rho_t\noise_t$) and the unbiasedness of the gradient estimator $\gradsvrg_t$. 
\end{proof}

\section{Ergodicity property of SGLD}



\subsubsection{Recurrence}



\begin{lemma}[Repeat of lemma \ref{lemma::recurrence}]
For a fixed $\delta>0$, let $n_0$ be the index such that $\eta_{n_0}\le \delta$, and $n_k$ be the sequence of iteration index $n_{k+1} = \min_s\{ s : s>n_k, \,  \eta_{n_k:s} \ge \delta   \}$. 

Under the regularization assumption \ref{assumption::regularization_conditions}, Lipschitz assumption \ref{as::grad_lipschitz_ld} and nonnegativity assumption \ref{assp::nonnegativity}, there exists a constant $C_1$ such that for $\alpha = 1-2\exp(-(1-C_1)\mu_1\delta)$, $B = 2( \psi_1 + \frac{2\eta_{n_0} L^3}{B_\e}\big( \mu_2 f(\x_0) + 2\psi_2 \big) + \frac{\rho_0^2 Ld}{2\eta_0})$ and $K = \frac{\ln\frac{f(\x_{n_0})}{\delta B}}{(1-C_1)\mu_1\delta}$, the stopping time sequence $\{\tau_k\}$, defined as $\tau_{0}=K$ and $\tau_{k+1} = \min\{t: t\ge \tau_k + 1, f(\x_{n_t})\le 2\delta B\}$ where $(\x_{t'})$ is the sequence generated by Algo.~\ref{alg::LD}, satisfies
\begin{equation}
    \E[\tau_j] \le \frac{4}{\alpha} + K + j\left(\frac{1}{2\alpha\delta }+1\right).
\end{equation}
\end{lemma}


\begin{proof}

Conditioned on $\mathcal{F}_{t}$ and $f(\widetilde{\x}^{(s)})<f(\x_0)$ for the largest $s$ such that $t\ge s B_\e$, we have 
\begin{align}
\label{ineqn::fcn_val_decrease}
\E[f(\x_{t+1})] &
\stackrel{\eqref{eq:descent}}{\le} f(\x_t) - \eta_t\gradnorm + \E \frac{\eta_t^2 L}{2}\|\widetilde{\nabla}_t\|^2 + \frac{\rho_t^2L}{2}\E\|\noise_t\|^2 \nonumber\\
&\stackrel{\eqref{ineqn::variance_gradsvrg_bound}}{\le}
f(\x_t) - \eta_t \gradnorm + \frac{\eta_t^2 L}{2} \left(2\gradnorm + 2 \frac{L^2}{B_\e}\|\x_t-\xbatch\|^2\right) + \frac{\rho_t^2 L d}{2} \nonumber\\
&= f(\x_t) - (\eta_t - \eta_t^2 L) \gradnorm + \frac{\eta_t^2 L^3}{B_\e}\|\x_t-\widetilde{\x}\|^2 + \frac{\rho_t^2 L d}{2} \nonumber \\ 
&\stackrel{\eqref{ineqn::ergo_regularisation_2}}{\le} f(\x_t) - (\eta_t - \eta_t^2 L) \gradnorm + \frac{2\eta_t^2 L^3}{B_\e}\big( \mu_2 (f(\x_t)+f(\x_0)) + 2\psi_2 \big) + \frac{\rho_t^2 L d}{2} \nonumber\\
&= (1+\frac{2\eta_t^2 L^3 \mu_2}{B_\e}) f(\x_t) - (\eta_t - \eta_t^2 L) \gradnorm + \frac{2\eta_t^2 L^3}{B_\e}\big( \mu_2 f(\x_0) + 2\psi_2 \big) + \frac{\rho_t^2 L d}{2} \nonumber\\
&\stackrel{\eqref{ineqn::ergo_regularisation_1}}{\le} (1+\frac{2\eta_t^2 L^3 \mu_2}{B_\e})f(\x_t) - (\eta_t - \eta_t^2 L) \big( \mu_1 f(\x_t) - \psi_1 \big) + \frac{2\eta_t^2 L^3}{B_\e} \big( \mu_2 f(\x_0) + 2\psi_2 \big) + \frac{\rho_t^2 L d}{2} \nonumber\\
&= \left(1 - \mu_1\eta_t + \eta_t^2 \left(\frac{2L^3\mu_2}{B_\e}+\mu_1 L\right) \right) f(\x_t) + (\eta_t - \eta_t^2 L)\psi_1 + \frac{2\eta_t^2 L^3}{B_\e} \big( \mu_2 f(\x_0) + 2\psi_2 \big) + \frac{\rho_t^2 L d}{2} \nonumber\\
&\le \mathrm{exp}\left( -(1-C_1)\mu_1\eta_t  \right) f(\x_t) + (\eta_t - \eta_t^2 L)\psi_1 + \frac{2\eta_t^2 L^3}{B_\e} \big( \mu_2 f(\x_0) + 2\psi_2 \big) + \frac{\rho_t^2 L d}{2}
\end{align}

Here $C_1$ is a positive constant such that $\eta_0(\frac{2L^3\mu_2}{\mu_1 B_\e}+ L)<C_1<1$ for small enough $\eta_0$.  

We introduce an index partition to characterize the function value decrease. Let $n_0$ be the index such that $\eta_{n_0}\le \delta$, and $n_k$ be the sequence of iteration index $n_{k+1} = \min_s\{ s : s>n_k, \,  \eta_{n_k:s} \ge \delta   \}$. Then $\eta_{n_k:n_{k+1}} \le 2 \delta$.

Before the proof proceeds, we recall the that the stepsize and variance are set (for some $\nu\ge 1$) as 
\begin{equation*}
\eta_t = \frac{\eta_0}{t^\nu} \textrm{ and } \rho_t = \frac{\rho_0}{t^{\nu/2}}.
\end{equation*}

Thus $\rho_t = \rho_0\sqrt{\frac{\eta_t}{\eta_0}}$.  Iterating \eqref{ineqn::fcn_val_decrease} $m$ times, we have
\begin{align}
\E f(\x_{t+m}) 
&\le \exp\left( -(1-C_1)\mu_1 \eta_{t:t+m-1} \right) f(\x_t) \; + \nonumber \\ 
&\quad\quad\sum_{i=t}^{t+m-1} \exp\left( -(1-C_1)\mu_1 \eta_{i+1:t+m-1} \right)\eta_i \left(\psi_1 + \frac{2\eta_i L^3}{B_\e}\big( \mu_2 f(\x_0) + 2\psi_2 \big) + \frac{\rho_0^2 Ld}{2\eta_0} \right)  \nonumber\\
&\le \exp\left( -(1-C_1)\mu_1 \eta_{t:t+m-1} \right) f(\x_t) + \sum_{i=t}^{t+m-1} \eta_i \left( \psi_1 + \frac{2\eta_i L^3}{B_\e}\big( \mu_2 f(\x_0) + 2\psi_2 \big) + \frac{\rho_0^2 Ld}{2\eta_0} \right)
\nonumber\\
&\le \exp\left( -(1-C_1)\mu_1 \eta_{t:t+m-1} \right) f(\x_t) +  \eta_{t:t+m-1} \left( \psi_1 + \frac{2\eta_t L^3}{B_\e}\big( \mu_2 f(\x_0) + 2\psi_2 \big) + \frac{\rho_0^2 Ld}{2\eta_0} \right) 
\label{ineqn::fcn_val_decrease_m}
\end{align}

Setting $t = n_{k-1}$ and $m = n_{k} - n_{k-1}$, inequality \eqref{ineqn::fcn_val_decrease_m} takes the form
\begin{align}
\E f(\x_{n_k}) &\le \exp\left( -(1-C_1)\mu_1 \eta_{n_{k-1}:n_k-1} \right) f(\x_{n_{k-1}}) +  \eta_{n_{k-1}:n_k-1} \left( \psi_1 +\frac{2\eta_{n_{k-1}} L^3}{B_\e}\big( \mu_2 f(\x_0) + 2\psi_2 \big) + \frac{\rho_0^2 Ld}{2\eta_0} \right) \nonumber\\
&\le \exp\left( -(1-C_1)\mu_1\delta \right) f(\x_{n_{k-1}}) +  \eta_{n_{k-1}:n_k-1} \left( \psi_1 + \frac{2\eta_{n_{k-1}} L^3}{B_\e}\big( \mu_2 f(\x_0) + 2\psi_2 \big) + \frac{\rho_0^2 Ld}{2\eta_0} \right) \label{ineqn:fcn_val_decrease_1_batch}\\
&\le \exp\left( -(1-C_1)k\mu_1\delta \right) f(\x_{n_0}) + \delta \underbrace{ 2\left( \psi_1 + \frac{2\eta_{n_0} L^3}{B_\e}\big( \mu_2 f(\x_0) + 2\psi_2 \big) + \frac{\rho_0^2 Ld}{2\eta_0} \right) }_{:= B}. 
\label{ineqn::fcn_val_decrease_batch}
\end{align}

Consider a function value threshold $M := 2\delta B$. From \eqref{ineqn::fcn_val_decrease_batch}, it follows that $\E f(\x_{n_k}) \le M$ when \[ k \ge \frac{\ln \frac{f(\x_{n_0})}{\delta B}}{(1-C_1)\mu_1 \delta}:= K\]

Now we show that the expected time for the function value to decrease to below this threshold $M$ is upper bounded by a finite number, thus justifying the recurrence of the iteration process to a compact sub-level set. (Note that under the regularization assumption \eqref{ineqn::ergo_regularisation_2}, all sub-level sets are compact.) To better exploit the indices partition $\{n_k\}$ of the iteration sequence, define $f(\x_{n_k}) := V_k$, and $\tau = \min\{k: k\ge K, f(\x_{n_k})\le M\}$. We claim that \[V_{\tau \wedge k} + \alpha\delta B \cdot (\tau\wedge k)\] is a supermartingale with $\alpha = 1 - 2\exp\left( -(1-C_1)\mu_1\delta \right)$, i.e.\ 
\begin{equation}
\label{defn::supermartingale}
\E[V_{\tau\wedge (k+1)} + \alpha\delta B(\tau\wedge (k+1)) | V_{\tau\wedge k}] \le V_{\tau\wedge k} + \alpha\delta B(\tau\wedge k)
\end{equation}

When $\tau \le k$, \eqref{defn::supermartingale} holds trivially. When $\tau \ge k+1$, then $V_{k+1} > M$. The relation \eqref{defn::supermartingale} to show in this case takes the form $\alpha \delta B \le V_k - \E[V_{k+1}|V_k]$. To let this happen, taking inequality \eqref{ineqn:fcn_val_decrease_1_batch} into consideration, a sufficient condition is $\E[V_{k+1}|V_k] \le \exp(-(1-C_1)\mu_1 \delta)V_k + \delta B \le V_k - \alpha\delta B$, i.e.\ $ (1+\alpha) \delta B \le \left(1-\exp(-(1-C_1)\mu_1 \delta) \right) V_k$. Considering that $\tau > k+1$ implies $V_k >M$, the previous sufficient condition to show can be further strengthened to $(1+\alpha)\delta B \le (1-\exp(-(1-C_1)\mu_1\delta) M$, which is catered for per definition of $\alpha$.  


To show that a sub-level set is going to be visited by the iteration sequence for infinitely many times, we introduce the stopping time sequence $\{\tau_k\}$ where $\tau_{0}=K$ and $\tau_{k+1} = \min\{t: t\ge \tau_k + 1, f(\x_{n_t})\le M\}$. 
Per the same argument as in previous paragraph, $\E\big[V_{\tau_{k+1}}+\alpha\delta B \tau_{k+1}\big|\tau_k\big] \le V_{\tau_k + 1}+\alpha\delta B (\tau_{k}+1)$, which gives
\[
\alpha\delta B \, \E[ \tau_{k+1} - \tau_k - 1 | \tau_k ] \le V_{\tau_k + 1} - \E[V_{\tau_{k+1}} | \tau_k]
\]
Taking total expectation, and summing over all $k$ from 0 to $j$ with $\tau_0 = K$, we have
\begin{equation}
    \alpha\delta B \, (\E[\tau_j] -K - j ) \le \sum_{k=0}^j \E[ V_{\tau_k + 1} - V_{\tau_{k+1}} ]
\end{equation}
By \eqref{ineqn::fcn_val_decrease}, $\E[V_{\tau_k + 1}] \le \exp\left( -(1-C_1)\mu_1 \eta_{\tau_k} \right) V_{\tau_k}+\frac{B}{2}\le V_{\tau_k}+\frac{B}{2}$, thus 
\[
\alpha\delta B \, (\E[\tau_j]-K-j) \le  \E[V_K - V_{\tau_j}] +j\frac{B}{2} \le 2M + j\frac{B}{2}
\]
i.e.\ 
\begin{equation*}
    \E[\tau_j] \le \frac{4}{\alpha} + K + j(\frac{1}{2\alpha\delta }+1)
\end{equation*}

\end{proof}

\subsubsection{Reachability}
The following Lemma \ref{lemma::variance_subset_selection} is stated as a fact, whose proof is straightforward computation, and will be needed for bounding the variance of the variance-reduced gradient estimator later in this section.

\begin{lemma}[Variance of subset selection]
\label{lemma::variance_subset_selection}
Consider a dataset $\{\mathbf{a}_i\}_{i=1}^N$ with mean  \[\bar{\mathbf{a}} = \frac{1}{N}\sum_{i=1}^N \mathbf{a}_i.
\]

Select $b$ elements uniformly ($1\le b \le N$) out of this dataset, and denote the index set of these selected elements as $\mathcal{I}$. The subsampled mean, which is a random variable,
is \[ \boldsymbol{\xi} =  \frac{1}{b}\sum_{i\in\mathcal{I}} \mathbf{a}_i.\] 

The variance of $\boldsymbol{\xi}$ is 
\begin{align}
\E_{\mathcal{I}} \| \boldsymbol{\xi} - \bar{\mathbf{a}} \|^2 &= \E_{\mathcal{I}} (\boldsymbol{\xi}^2 - 2\<\boldsymbol{\xi}, \bar{\mathbf{a}}\> + \bar{\mathbf{a}}^2) = \E_{\mathcal{I}} \boldsymbol{\xi}^2 - \bar{\mathbf{a}}^2 \nonumber\\
&= \frac{N-b}{N^2 b} \mathrm{Var}[\mathbf{a}] = \frac{N-b}{(N-1) b} \left( \frac{1}{N}\sum_{i=1}^N \| \mathbf{a}_i - \bar{\mathbf{a}} \|^2 \right)
\label{eqn::subset_selection_variance}
\end{align}
\end{lemma}

Lemma \ref{lemma::y_k_bound} will be used to show that inside a stepsize batch, the reachability property will not be hindered by the gradient descent part in the iteration, thus allowing the Gaussian noise terms to give the desired property.

\begin{lemma}
\label{lemma::y_k_bound}
Let $n$ be a positive
integer, then for any sequence $a_k > 0$ such that there is a constant $\nu$ and $\sum_{j=1}^n a_j \le 2\nu$,  let $\mathcal{F}_\z$ denote the $\sigma$-algebra generated by $\z_1,\cdots,\z_n$. Suppose $\boldsymbol{\xi}_k$ is a sequence of random vectors such that 
\[
\E(\boldsymbol{\xi}_k\,|\, \mathcal{F}_\z) = 0\, \quad \E(\|\boldsymbol{\xi}_k\|^2\,|\, \mathcal{F}_\z) \le C_2
\]
Let $\y_k = \sum_{j=1}^k a_j \boldsymbol{\xi}_j$, then 
\begin{equation}
    \pr(\|\y_k\| \le 4\nu\sqrt{C_2}) \ge \frac{1}{2}
\end{equation}
\end{lemma}

\begin{proof}
In the proof for this lemma, all expectations are conditioned on $\mathcal{F}_\z$. 
With Jensen's inequality, $(\E\|\boldsymbol{\xi}_k\|)^2 \le \E(\|\boldsymbol{\xi}_k\|^2) \le C_2$. By Markov's inequality,
\begin{align*}
    \pr(\|\y_k\|\ge 4\nu \sqrt{C_2})\le \frac{\E \|\y_k\|}{4\nu\sqrt{C}_2}  = \frac{\E \|\sum_{j=1}^k a_j \boldsymbol{\xi_j}\|}{4\nu \sqrt{C}_2} \le \frac{\E \sum_{j=1}^k a_j \|\boldsymbol{\xi}_j\|}{4\nu\sqrt{C_2}}  \le \frac{1}{2}
\end{align*}
\end{proof}


\begin{lemma}[Repeat of lemma \ref{lemma::brownian_ergodicity}]
Given any sequence $a_k > 0$,  
let $\z_k = \sum_{i=1}^k \rho_0\sqrt{a_i}\noise_i$ where $\noise_i \sim\mathcal{N}(0,I_d)$, for any target vector $\z^\star$ and distance $r$, there exists a non-negative function $p_1$ such that  
\[
\pr(\|\z_n-\z^\star\|\le r, \, \|\z_k\| \le \|\z^\star\|+r \,\, \forall k=1,\cdots, n )\ge p_1(r,\rho_0,t_n,\z^\star)
\]
with $p_1(0,\rho_0,t_n,\z^\star) = p_1(r,0,t_n,\z^\star) = 0$. 
\end{lemma}
\begin{proof}
We first give lower bounds to factors $\pr(\|\z_n-\z^\star\|\le r )$ and $\pr( \|\z_k\| \le \|\z\|+r \,\, \forall k=1,\cdots, n )$ respectively, and then conclude the proof with $\pr(\|\z_n-\z^\star\|\le r, \, \|\z_k\| \le \|\z\|+r \,\, \forall k=1,\cdots, n ) \ge \pr(\|\z_n-\z^\star\|\le r ) \cdot \pr( \|\z_k\| \le \|\z\|+r \,\, \forall k=1,\cdots, n )$.


For $\pr(\|\z_n-\z^\star\|_2 \le r )$, $ \|\z_n - \z^\star\|_2 \le \|\z_n - \z^\star\|_1 = \sum_{\mathrm{dim}=1}^d |(\z_n)_{\mathrm{dim}} - z^\star_{\mathrm{dim}} | $, therefore $\pr(\|\z_n-\z^\star\|_2 \le r ) \ge \pr( |(\z_n)_{\mathrm{dim}} - z^\star_{\mathrm{dim}} | \le \frac{r}{d}\,\,\forall \mathrm{dim}\in[d] ) = \prod_{\mathrm{dim}=1}^d \pr(|(\z_n)_{\mathrm{dim}} - z^\star_{\mathrm{dim}} | \le \frac{r}{d})$. 
Notice that $(\z_n)_{\mathrm{dim}}$ has the distribution of the Brownian motion $B_{t_n}$ where $t_k = \rho_0^2\sum_{i=1}^k a_i$, $k=1,2,\cdots,n$. By \cite{Karlin_Taylor_BM_book_Ch7}, $\dis \pr(|(\z_n)_{\mathrm{dim}} - z^\star_{\mathrm{dim}} | \le \frac{r}{d}) = \int_{\max\{z^\star_{\mathrm{dim}} - \frac{r}{d},0\}}^{z^\star_{\mathrm{dim}} + \frac{r}{d}} p_{t_n} (z_{\mathrm{dim}}^\star,y)\,\mathrm{d} y $, where $p_{t} (x,y) = \sqrt{\frac{2}{\pi t}}\exp(-\frac{x^2+y^2}{2t})\cosh(\frac{xy}{t}) $. Hence,
\begin{equation}
\label{ineqn::mbm_bd_part1}
\pr(\|\z_n-\z^\star\|_2 \le r ) 
\ge  \left( \min_{\mathrm{dim}} \int_{\max\{z^\star_{\mathrm{dim}} - \frac{r}{d},0\}}^{z^\star_{\mathrm{dim}} + \frac{r}{d}} p_{t_n} (z_{\mathrm{dim}}^\star,y)\,\mathrm{d} y  \right)^d
\end{equation}

For $\pr(\|\z_k\|\le \|\z^\star\|+r,\, \forall k\in[n])$, 
we have the following lower bound:
\begin{align*}
    \pr(\|\z_k\|\le \|\z^\star\|+r, \, \forall k\in[n])
    &\ge \pr\left(\max_k |(\z_k)_\textrm{dim}| \le \frac{\|\z^\star\|+r}{\sqrt{d}},\, \forall \textrm{dim}\in[d]\right)\\
    &= \pr\left(\max_k |\underbrace{(\z_k)_1}_{\textrm{1D B.M.}}| \le \frac{\|\z_\star\|+r}{\sqrt{d}}\right)^d \\
    &= \left(1-\pr\left(\max_k |(\z_k)_1| \ge \frac{\|\z^\star\|+r}{\sqrt{d}}\right)\right)^d
\end{align*}
Now notice $\pr(\max_k|(\z_k)_1|\ge \frac{\|\z^\star\|+r}{\sqrt{d}}) = \pr\bigg(\max_k (\z_k)_1 > \frac{\|\z^\star\|+r}{\sqrt{d}} \textrm{ or } \min_k (\z_k)_1 < -\frac{\|\z^\star\|+r}{\sqrt{d}} \bigg) \le \pr \big( \max_k (\z_k)_1 > \frac{\|\z^\star\|+r}{\sqrt{d}} \big) + \pr \big( \min_k (\z_k)_1 < -\frac{\|\z^\star\|+r}{\sqrt{d}} \big)  = 2 \pr \big( \max_k (\z_k)_1 > \frac{\|\z^\star\|+r}{\sqrt{d}} \big) $. Then, by the reflection principle of Brownian motion, 
\[
\pr\left(\max_k (\z_k)_1 > \frac{\|\z^\star\|+r}{\sqrt{d}} \right) = 2 \pr\left( (\z_n)_1 \ge \frac{\|\z^\star\|+r}{\sqrt{d}} \right)
\]
Therefore,
\begin{align}
\label{ineqn::mbm_bd_part2}
    \pr(\|\z_k\|\le \|\z^\star\|+r, \, \forall k\in[n]) 
    &\ge \left(1- 4 \pr\left(\underbrace{(\z_n)_1}_{\sim\mathcal{N}(0,t_n)}\ge \frac{\|\z^\star\|+r}{\sqrt{d}}\right) \right)^d \nonumber\\
    &= \left(1- 4 \pr\left( \frac{(\z_n)_1}{\sqrt{t_n}}\ge \frac{\|\z^\star\|+r}{\sqrt{d t_n}} \right) \right)^d \nonumber\\
    &= \left(1- 2 \pr\left( \frac{(\z_n)_1}{\sqrt{t_n}}\ge \frac{\|\z^\star\|+r}{\sqrt{d t_n}} \right) - 2\pr\left( \frac{(\z_n)_1}{\sqrt{t_n}}\le -\frac{\|\z^\star\|+r}{\sqrt{d t_n}} \right) \right)^d \nonumber\\
    &=\left(1-2 \big( 1-\pr\left( -\frac{\|\z^\star\|+r}{\sqrt{d t_n}} \le\frac{(\z_n)_1}{\sqrt{t_n}}\le \frac{\|\z^\star\|+r}{\sqrt{d t_n}} \right) \big) \right)^d\nonumber\\
    &=\left(2\int_{-\frac{\|\z^\star\|+r}{\sqrt{d t_n}}}^{\frac{\|\z^\star\|+r}{\sqrt{d t_n}}} \frac{\exp(-x^2/2)}{\sqrt{2\pi}}\,\mathrm{d}x -1 \right)^d \nonumber\\
    &\ge \left( 4 \frac{\|\z^\star\|+r}{\sqrt{2\pi d t_n}} \exp\left( -\frac{1}{2}\frac{(\|\z^\star\|+r)^2}{d t_n} \right) -1 \right)^d
\end{align}


Let $p_1(r,\rho_0,t_n,\z^\star)$ be the product of two lower bounds \eqref{ineqn::mbm_bd_part1} and \eqref{ineqn::mbm_bd_part2} above, recall that $t_n = \rho_0^2 \sum_{i=1}^n  a_i $, we define $p_1(r,\rho_0,t_n,\z^\star)$ as the following,

\begin{equation}
    p_1(r,\rho_0,t_n,\z^\star) :=  \left( \min_{\mathrm{dim}} \int_{\max\{z^\star_{\mathrm{dim}} - \frac{r}{d},0\}}^{z^\star_{\mathrm{dim}} + \frac{r}{d}} p_{t_n} (z_{\mathrm{dim}}^\star,y)\,\mathrm{d} y  \right)^d \cdot \left( 4 \frac{\|\z^\star\|+r}{\sqrt{2\pi d t_n}} \exp\left( -\frac{1}{2}\frac{(\|\z^\star\|+r)^2}{d t_n} \right) -1 \right)^d
\end{equation}
where $p_{t_n} (x,y) = \sqrt{\frac{2}{\pi t}}\exp(-\frac{x^2+y^2}{2t_n})\cosh(\frac{xy}{t_n}) $.

To make the dependence of the first factor in $p_1$ on parameters more explicit, for some $\xi\in(\max\{z^\star_{\mathrm{dim}} - \frac{r}{d},0\},z^\star_{\mathrm{dim}} + \frac{r}{d})$, \begin{align*} \int_{\max\{z^\star_{\mathrm{dim}} - \frac{r}{d},0\}}^{z^\star_{\mathrm{dim}} + \frac{r}{d}} p_{t_n} (z_{\mathrm{dim}}^\star,y)\,\mathrm{d} y  
&\ge  \sqrt{\frac{2}{\pi t_n}} \exp(-\frac{ (z_{\textrm{dim}}^\star)^2 +  \xi^2}{2t_n})\cosh(\frac{z_{\textrm{dim}}^\star \xi}{t_n})  \frac{r}{d}\\
&= \sqrt{\frac{2}{\pi t_n}} \left(\exp( -\frac{(z^\star_{\mathrm{dim}}-\xi)^2}{2t_n} ) + \exp( -\frac{(z^\star_{\mathrm{dim}}+\xi)^2}{2 t_n} ) \right)  \frac{r}{2d}\\
&\ge \sqrt{\frac{2}{\pi t_n}} 2\sqrt{ \exp( -\frac{(z^\star_{\mathrm{dim}}-\xi)^2}{2t_n}  -\frac{(z^\star_{\mathrm{dim}}+\xi)^2}{2 t_n} )  } \frac{r}{2d}\\
&= \sqrt{\frac{2}{\pi t_n}} \exp( -\frac{(z^\star_{\mathrm{dim}})^2 + \xi^2}{2t_n}  )   \frac{r}{d}\\
&\ge \sqrt{\frac{2}{\pi t_n}} \exp( -\frac{(z^\star_{\mathrm{dim}})^2 + (z^\star_{\mathrm{dim}}+\frac{r}{d})^2}{2t_n}  )   \frac{r}{d}
\end{align*}

We thus redefine $p_1$ as
\begin{equation}
    \label{defn::p1}
    p_1(r,\rho_0,t_n,\z^\star) := \left( \min_{\mathrm{dim}} \sqrt{\frac{2}{\pi t_n}} \exp( -\frac{(z^\star_{\mathrm{dim}})^2 + (z^\star_{\mathrm{dim}}+\frac{r}{d})^2}{2t_n}  ) \frac{r}{d}  \right)^d \cdot \left( 4 \frac{\|\z^\star\|+r}{\sqrt{2\pi d t_n}} \exp\left( -\frac{1}{2}\frac{(\|\z^\star\|+r)^2}{d t_n} \right) -1 \right)^d
\end{equation}
then we have the lemma \ref{lemma::brownian_ergodicity}.
\end{proof}


\begin{lemma}[Repeat of lemma \ref{lemma::reachability}]
Assume the same stepsize batch setting as in lemma \ref{lemma::recurrence} and the gradient Lipschitz condition in assumption \ref{as::grad_lipschitz_ld}, with respect to an arbitrary target point $\mathbf{s}$ in the level set $\{\x: f(\x)<\mathcal{O}(\widetilde{\eps})\}$, there is a constant $C_{21} \propto dL$ and a constant $C_{22}$, for any $\eps>0$, there exists a non-negative $p_1(\widetilde{\eps},\rho_0,t_{n_{i+1}},\mathbf{s})$ such that
\begin{equation}
    \pr\left( \| \x_{n_{i+1}} - \mathbf{s} \| \le \widetilde{\eps} \right) > \frac{1}{2}p_1(\widetilde{\eps},\rho_0,t_{n_{i+1}},\mathbf{s})
\end{equation}
where $\widetilde{\eps} = \eps + \delta C_{21} + 2\delta \sqrt{C_{22}} $, and $p_1$ is given in lemma \ref{lemma::brownian_ergodicity}.
\end{lemma}

\begin{proof}

Denote $\x_o=\x_{n_i}$ and $\mathbf{d} =  \mathbf{s}- \x_o$. Recall the SGLD scheme in a batch goes as
\begin{align}
\label{eqn::reachability_procession}
\x_{k+1} &= \x_k - \eta_k \gradsvrg_k + \rho_k \noise_k = \x_o - \sum_{l=0}^k \eta_l\gradsvrg_l + \rho_l\noise_l \nonumber\\
& = \x_o - \sum_{l=0}^k \eta_{l+o}\left( \nabla f(\x_{l+o}) + \big( \frac{1}{B} \nabla f_{I_{l+o}}(\x_{l+o}) - \nabla f(\x_l) \big) - \big( \frac{1}{B} \nabla f_{I_{l+o}}(\widetilde{\x}) - \nabla f(\widetilde{\x}) \big)  \right) + \sum_{l=0}^k \rho_{l+o}\noise_{l+o} \nonumber\\
&= \x_o - \sum_{l=0}^k \eta_{l+o}\nabla f(\x_{l+o}) - \y_k + \z_k
\end{align}
where we define $\y_k := \sum_{l=0}^k  \eta_l\big( \frac{1}{B} \nabla f_{I_l}(\x_l) - \nabla f(\x_l) \big) - \eta_l\big( \frac{1}{B} \nabla f_{I_l}(\widetilde{\x}) - \nabla f(\widetilde{\x}) \big) $ and $\z_k := \sum_{l=0}^k \rho_0 \sqrt{\frac{\eta_l}{\eta_0} }\noise_l$. Note that $\widetilde{\x}$ can change as moving from stepsize batch $i$ to $i+1$ may involve different SVRG batch reference points. 


Let $m = n_{i+1} - n_i$. Per law of total probability, Denote the event $\mathcal{E}_{i+1} = \{\|\z_{n_{i+1}-1}-\mathbf{d}\| \le \eps, \, \|\z_k\|\le \|\mathbf{d}\| + \eps \,\,\forall k\in[m]+n_i\}$, 
then \begin{equation}
\pr\left( \| \x_{n_{i+1}} - \mathbf{s} \| \le \widetilde{\eps} \right) 
\ge \pr\left(  \| \x_{n_{i+1}} - \mathbf{s} \| \le \widetilde{\eps} \,\big| \, \mathcal{E}_{i+1} \right) \cdot \pr\left(\mathcal{E}_{i+1} \right)
\end{equation}
where $\pr\left(\mathcal{E}_{i+1} \right)$ is lowered bounded by $p_1$ from lemma \ref{lemma::brownian_ergodicity}. What is left in this proof is to bound the first factor in the above equation.

Now we bound the 
gradient difference terms in \eqref{eqn::reachability_procession}, thus computing the probability $ \pr\left(  \| \x_{n_{i+1}} - \mathbf{s} \| \le \widetilde{\eps} \,\big| \, \mathcal{E}_{i+1} \right)$. 


As the gradient is bounded inside a stepsize batch $\{\eta_{n_i}:\eta_{n_{i+1}-1}\}$, by lemma \ref{lemma::variance_subset_selection}, each summation term in $\y_k$ has the following variance upper bound \[\frac{n-B}{(n-1)B} \frac{1}{n}\sum_{i=1}^n \|\nabla f_i(\x_l) - \nabla f(\x_l)\|^2 \le \frac{n-B}{(n-1)B} \frac{1}{n} \max_{l\in[m]+o} \sum_{i=1}^n \|\nabla f_i(\x_l) - \nabla f(\x_l)\|^2 :=C_{22}\] 

Within the gradient batch, the term $\|\sum_{l=0}^{n_{i+1}-1} \eta_{l+o} \nabla f(\x_{l+o}) \| \le \max_{l\in[n_{i+1}-1]} \|\nabla f(\x_{l+o})\| \sum_{l=0}^{n_{i+1}-1} \eta_{l+o}  \le \delta \max_{l\in[n_{i+1}-1]} \|\nabla f(\x_{l+o})\|$. After the iteration has proceeded sufficiently, per convergence properties first-order stationary points in lemma \ref{lemma::lyapunov_bound_gradient}, the boundedness of gradients in the gradient batch can be controlled by a constant $C_{21} = \mathcal{O}(dL)$.     
The unbiasedness of SVRG gradient estimator makes lemma \ref{lemma::y_k_bound} applicable to $\y_k$:
\begin{align*}
    \pr( \|\x_{n_{i+1}}-\mathbf{s}\|\le \widetilde{\eps} \,|\, \mathcal{E}_{i+1} ) &= \pr( \|\x_o - \sum_{l=0}^{n_{i+1}-1} \eta_{l+o}\nabla f(\x_{l+o}) - \y_{n_{i+1}-1} + \z_{n_{i+1}-1} -\mathbf{s}\|\le \widetilde{\eps} \,|\, \mathcal{E}_{i+1} )\\
    &= \pr( \| - \sum_{l=0}^{n_{i+1}-1} \eta_{l+o}\nabla f(\x_{l+o}) - \y_{n_{i+1}-1} + \z_{n_{i+1}-1} -\mathbf{d}\|\le \widetilde{\eps} \,|\, \mathcal{E}_{i+1} )\\
    &\ge \pr(\|\sum_{l=0}^{n_{i+1}-1} \eta_{l+o}\nabla f(\x_{l+o})\| \le \delta C_{21} , \|\y_{n_{i+1}-1}\|\le 4\delta C_{22},  \,|\,\mathcal{E}_{i+1} )\\
    &\ge \pr(\|\y_{n_{i+1}-1}\|\le 4\delta C_{22},  \,|\,\mathcal{E}_{i+1} ) \ge \frac{1}{2}
\end{align*}
where the last inequality is due to the fact that the first part of the event is a certain event with the iteration having proceeded sufficiently and the choice of $\delta$ to bound the sum of stepsize update in a batch.

\end{proof}

Now we are ready to prove the ergodicity result for the SGLD-VR scheme with the recurrence and reachability results above.
\begin{theorem}[Repeat of theorem \ref{thm::ergodicity}]
Under regularization condition \ref{assumption::regularization_conditions}, Lipschitz assumption \ref{as::grad_lipschitz_ld} and nonnegativity assumption \ref{assp::nonnegativity}, with the same parameter setting as in lemma \ref{lemma::recurrence}, for any $\widetilde{\eps}>0$, $p>0$ and a point $\mathbf{s}$ which locates in the level set $\{\x: f(\x)\le \mathcal{O}(\widetilde{\eps})\}$, there is a \begin{equation}
    T = \mathcal{O}\left(\frac{1}{p\mu_1\big(4\eta_0L^3 \frac{\mu_2 f(\x_0) + 2\psi_2}{B_\e} + \frac{\rho_0^2}{\eta_0} Ld\big)}\left(1 + \ln f(\x_0)   + \frac{(d\|\mathbf{s}\| + \widetilde{\varepsilon}) ^d }{ \big( (\frac{4}{\sqrt{2\pi}}-1)\mathrm{e}^{-1/2} \widetilde{\varepsilon} \big)^d}  \right) \right)
\end{equation} 
such that
\begin{equation}
    \pr(\|\x_t - \mathbf{s}\|\le \widetilde{\eps} \textrm{ for some } t<T) \ge 1-p
\end{equation}
\end{theorem}

\begin{proof}[Proof of Thm.~\ref{thm::ergodicity}]

Recall the definition of the stopping time sequence: $\tau_0 = K$, $\tau_{t+1} = \min\{t: t\ge \tau_k + 1, f(\x_{n_t}) \le M = 2\delta B\}$. By defining $\tau_* = \min\{t: t>0, \|\x_{n_t} - \mathbf{s}\| \le \widetilde{\eps}\}$ and setting $\delta = \frac{\widetilde{\eps}}{2B}$, we show that $\pr(\tau_* \ge T) \le \widetilde{p}$ with a proper choice of $T$. For any $J$,
\begin{align*}
    \pr(\tau_* \ge T) &= \pr(\tau_* \ge T, \tau_J >T) + \pr(\tau_* \ge T, \tau_J <T)\\
    &\le \pr( \tau_J>T ) + \pr( \|\x_{n_{\tau_k}+1}-\mathbf{s}\|> \widetilde{\eps}, \tau_J \le T, \,\,\forall k \in[J] )\\
    &\le \pr( \tau_J>T ) + \pr( \|\x_{n_{\tau_k}+1}-\mathbf{s}\|> \widetilde{\eps}, \,\,\forall k \in[J] )
\end{align*}
Lemma \ref{lemma::recurrence} gives that $\E\tau_J \le  \frac{4}{\alpha} + K + J(\frac{1}{2\alpha\delta }+1)$, thus by Markov inequality 
\begin{equation}
    \label{ineqn::bdd_first_term_p2}
\pr(\tau_J > T) \le \frac{\E[\tau_J]}{T} \le \frac{\frac{4}{\alpha} + K + J(\frac{1}{2\alpha\delta }+1)}{T}
\end{equation}
To ensure the last bound is below the pre-specified threshold $\dis \frac{1}{2}p$, we need to take 
\begin{equation}
\label{eqn::T1}
    T = \left[\frac{\frac{4}{\alpha} + K + J(\frac{1}{2\alpha\delta }+1)}{\frac{p}{2}}\right] + 1
\end{equation}

By lemma \ref{lemma::reachability}, there is a $p_2 = p_1/2>0$ such that 
\begin{equation}
    \label{ineqn::bdd_second_term_p2}
\pr(\|\x_{n_{\tau_k}+1}-\mathbf{s}\|>\widetilde{\eps}, \,\,\forall k \in[J] ) = \prod_{k=1}^J \pr(\|\x_{n_{\tau_k}+1}-\mathbf{s}\|>\widetilde{\eps} )\le (1-p_2)^J \stackrel{\textrm{let}}{\le} \frac{p}{2}
\end{equation}
To ensure the upper bound to be less than $\frac{p}{2}$, a sufficient condition is that $\dis J>\frac{\ln \frac{p}{2}}{\ln ( 1-p_2(\widetilde{\varepsilon},\rho_0,t_{n_{\tau_J}+1}) )} > \frac{\ln \frac{2}{p}}{p_2(\widetilde{\varepsilon},\rho_0,t_{n_{\tau_J}+1})} $. 

We consider the dependence of $T$ on the error tolerance $\widetilde{\varepsilon}$, dimension $d$ and initial perturbation parameter $\rho_0$. Recall $p_2= \frac{1}{2}p_1$ and we will upper bound it to rid of the dependence on $t_n$: 
\begin{align*}
&p_2(\widetilde{\varepsilon},\rho_0,t_n,\mathbf{s})  \\
&= \frac{1}{2} \left( \min_{\mathrm{dim}} \sqrt{\frac{2}{\pi t_n}} \exp( -\frac{(s_{\mathrm{dim}})^2 + (s_{\mathrm{dim}}+\frac{\widetilde{\varepsilon}}{d})^2}{2t_n}  ) \frac{\widetilde{\varepsilon}}{d}  \right)^d \cdot \left( 4 \frac{\|\mathbf{s}\|+\widetilde{\varepsilon}}{\sqrt{2\pi d t_n}} \exp\left( -\frac{1}{2}\frac{(\|\mathbf{s}\|+\widetilde{\varepsilon})^2}{d t_n} \right) -1 \right)^d\\
&\le \frac{1}{2} \left( \min_{\mathrm{dim}} \sqrt{\frac{2}{\pi \left( (s_{\mathrm{dim}})^2 + (s_{\mathrm{dim}}+\frac{\widetilde{\varepsilon}}{d})^2 \right) }} \exp( -\frac{1}{2}  ) \frac{\widetilde{\varepsilon}}{d}  \right)^d 
\cdot 
\left( 4\frac{1}{\sqrt{2\pi}} -1 \right)^d
\end{align*}
Therefore, a sufficient condition for \eqref{ineqn::bdd_second_term_p2} to hold is
\begin{equation}
\label{eqn::T2}
    J > \left(\ln\frac{2}{p}\right) \max_{\mathrm{dim}} \frac{2}{ \left(  \sqrt{\frac{2}{\pi \left( (s_{\mathrm{dim}})^2 + (s_{\mathrm{dim}}+\frac{\widetilde{\varepsilon}}{d})^2 \right) }} \exp( -\frac{1}{2}  ) \frac{\widetilde{\varepsilon}}{d}  \right)^d 
\cdot 
\left( 4\frac{1}{\sqrt{2\pi}} -1 \right)^d }
\end{equation}

Recall from lemma \ref{lemma::recurrence} parameter settings $K =\frac{\ln \frac{f(\x_{n_0})}{\delta B}}{(1-C_1)\mu_1 \delta}$, $B = 2( \psi_1 + \frac{2\eta_{n_0} L^3}{B_\e}\big( \mu_2 f(\x_0) + 2\psi_2 \big) + \frac{\rho_0^2 Ld}{2\eta_0}) $ and $\alpha = 1-2\exp(-(1-C_1)\mu_1\delta)$.
In light of the remark post the lemma \ref{lemma::recurrence}, $\delta$ is to be set as $\delta\propto \widetilde{\eps}B^{-1}$ for the purpose of minimizing the empirical risk. Combining \eqref{eqn::T1} and \eqref{eqn::T2}, the total amount of time needed for \eqref{ineqn::ergodicity_main_result} to hold is 
\begin{align}
    T&=  \left[\frac{\frac{4}{\alpha} + K + J(\frac{1}{2\alpha\delta }+1)}{\frac{p}{2}}\right] + 1 \nonumber\\
    &= \widetilde{\mathcal{O}}\left(\frac{1 + \ln f(\x_0)   + \frac{(d\|\mathbf{s}\| + \widetilde{\varepsilon}) ^d }{ \widetilde{\eps}\big( (\frac{4}{\sqrt{2\pi}}-1)\mathrm{e}^{-1/2} \widetilde{\varepsilon} \big)^d}}{\widetilde{\eps} p\mu_1\big(\psi_1+ 2\eta_0L^3 \frac{\mu_2 f(\x_0) + 2\psi_2}{B_\e} + \frac{\rho_0^2}{\eta_0} Ld\big)} \right)
\end{align}
\end{proof}

%% file: second_order_convergence.tex
\subsection{Proofs of second-order stationary point convergence property}

\begin{proof}[Proof of Thm.~\ref{thm::second_order_convergence}]

~\paragraph{Step 1} 
Assume the stepsize decay parameter $\nu\in[1,2]$ for simplicity. 
We show that $\Delta_i  := \sum_{l=n_{i}}^{n_{i+1}-1} \sqrt{\eta_i} \noise_i$ will lead to saddle point escape, i.e. $\Delta_i^\trans \nabla^2 f(\x_{\textrm{fsp}}) \Delta_i \le -\zeta $. Specifically, we show that the projection of $\Delta_i$  on the direction of $\lambda_{\mathrm{min}}$ is than $\zeta $ with high probability, which exploits the property of Brownian motion and the idea that the trapping region is thin when faced with LD \cite{Huang_PPD}.

At a fixed first-order stationary point $\fsp$, due to the spatial homogeneity of Brownian motion, w.l.o.g.\, assume that $\mathbf{e}_1$ is the unit eigenvector corresponding to the smallest eigenvalue of $\nabla^2 f(\fsp)$. To have $\Delta_i^\trans \nabla^2 f(\x_{\textrm{fsp}}) \Delta_i \le -\zeta$, a sufficient condition is $\lambda_{\min}(\nabla^2 f(\fsp))(\Delta_i)_1 ^2 + L (\|\Delta_i\|^2 - (\Delta_i)_1 ^2) \le -\zeta $. Assume for now that $\|\Delta_i\|^2 \le r^2$, then this condition can be phrased as $\lambda_{\min}(\nabla^2 f(\fsp))(\Delta_i)_1 ^2 + L (r^2 - (\Delta_i)_1 ^2) \le -q(\Delta_i)_1 ^2 + L (r^2 - (\Delta_i)_1 ^2)\le -\zeta $, i.e.\ the first component of $\Delta_i$ satisfies
\begin{equation}
\label{ineqn::first_dim_projection_requirement}
(\Delta_i)_1^2 \ge \frac{\zeta + Lr^2}{L +q } := Q
\end{equation}
Now we compute the probability for \eqref{ineqn::first_dim_projection_requirement} to fail within the time $T_i := \sum_{l=n_{i}}^{n_{i+1}-1} \sqrt{\eta_l} $ for a standard 1D Brownian motion. Define $\tau_{Q} = \min\{ t\,|\, \big((\Delta_i)_1(t)\big)^2 \ge Q\}$. Then 
\[ 
\pr( \eqref{ineqn::first_dim_projection_requirement}\textrm{ fails to hold within time } T_i ) = \pr(\tau_Q > T_i) \le \frac{\E \tau_Q}{T_i} = \frac{Q}{d T_i} .
\] 
Here we point out that the failure probability for \eqref{ineqn::first_dim_projection_requirement} is low due to the large denominator. 
$T_i = \sum_{l=n_i}^{n_{i+1}-1}\sqrt{\eta_l} \le \sqrt{(n_{i+1}-n_i) \sum_{l=n_i}^{n_{i+1}-1} \eta_l} \approx \sqrt{(n_{i+1}-n_i)\delta}$. Note that $\displaystyle n_{i+1} - n_i = \mathcal{O}\left( n_i\exp(\delta) \right)$, then $n_i = \mathcal{O}(\exp(i\delta))$, thus \begin{equation}
\label{eqn::T_i_stepsize_batch}
    T_i = \exp(\mathcal{O}(i\delta))
\end{equation}

\textbf{Remark}: consider the case $\nu = 1$ as an example for the preceding claim. As $\sum_{l=n_i}^{n_{i+1}-1} \eta_l \approx \delta$ and $n_0 = 1$, $n_i \approx \exp(i\delta)$ and $n_{i+1} = n_i \exp(\delta)$. The corresponding time in continuous domain 
\begin{align*}
T_i 
&= \sum_{l=n_i}^{n_{i+1}-1}\sqrt{\eta_l} \approx \int_{n_i}^{n_{i+1}-1} \sqrt{\eta_0} \frac{1}{\sqrt{t}}\,\mathrm{d}t = 2\sqrt{\eta_0}(\sqrt{n_{i+1}-1} - \sqrt{n_i})\\
&\approx 2\sqrt{\eta_0}(\sqrt{n_{i+1}} - \sqrt{n_i}) = 2\sqrt{\eta_0n_i}(\sqrt{\exp(\delta)}-1) = 2\sqrt{\eta_0\exp(i\delta)}(\sqrt{\exp(\delta)}-1).
\end{align*}

\paragraph{Step 2} 
We show that when $\x \in \mathcal{U}(\x_{\mathrm{fsp}},r)$ where $\|\Delta_j\|<r$ for $j=n_i, n_i+1,\cdots, n_{i+1}-1$, $\|\nabla f(\x)\|<\eps$, thus the first order expansion does not contribute to function value change. Due to the Lipschitz gradient, set 
\[ 
r = \max\left\{ \frac{\eps}{L}, \sqrt{\frac{3}{Lq}}\eps \right \}.
\] 

While the projection onto $\mathbf{e}_1$ builds up, we compute the probability that the iteration is still constrained within the $\eps$-neighborhood of $\fsp$: 
\begin{align}
\pr( \Delta_i^2 - (\Delta_i)_1^2 \le r^2-Q \textrm{ when } t \le T_i) &= \pr(\sum_{i=2}^d \hat{x}_i^2 \le r^2-Q \textrm{ when } t \le T_i) \nonumber \\
&= \int_0^{\sqrt{r^2-Q}}  (T_i)^{-\frac{d-1}{2}} 
\frac{1}{\Gamma(\frac{d-2}{2})}\exp(-\frac{y^2}{2T_i})  y^{d-2}\,\mathrm{d} y \nonumber\\
&\approx (T_i)^{-\frac{d-1}{2}} 
\frac{1}{\Gamma(\frac{d-2}{2})} \int_0^{\sqrt{r^2-Q}} y^{d-2}\,\mathrm{d} y \nonumber\\
&= (T_i)^{-\frac{d-1}{2}} 
\frac{1}{\Gamma(\frac{d-2}{2})}(r^2-Q)^{\frac{d-1}{2}}
:= P_i
\label{defn::ssp_constraint_prob}
\end{align}

As $T_i$ increases exponentially w.r.t.\ index $i$, $P_i$ decreases accordingly, i.e.\ , within a stepsize batch, the probability for the iteration to remain bounded within the vicinity of a FSP is decreasing. Hence, the saddle point escape process can be thought of as a binomial trial with decreasing success probability, and the expected time for the iteration process to escape all saddle points is at least proportional to $\Gamma(\frac{d-2}{2})$. 


\paragraph{Step 3} 
Show that the update $\x' = \x + \Delta_i$ will lead to function value decrease, thus the SGLD algorithm has to terminate, thus converging to SSP.

Denote the event $\mathcal{A}_i=\{ \Delta_i^\trans \nabla^2 f({\fsp}_i) \Delta_i \le -\zeta \textrm{ and } \|\Delta_i\|\le r \}$. From steps 1 and 2, $\pr(\mathcal{A}_i) \ge (1-\frac{Q}{d T_i}) P_i$. We show that under the assumption that event $\mathcal{A}_i$ happens, function value decrease is guaranteed. 
Note that within a minibatch, 
\begin{align} 
\E\|\x_t-\xbatch\|^2 &= \E\left\|\sum_{u=o}^{B_\e-1} \x_{u+1} - \x_u \right\|^2 = \E\left\|\sum_{u=o}^{B_\e-1} \eta_u \big( \nabla f_{i_{u}}(\x_{u+1}) - \nabla f_{i_u} (\xbatch) + \nabla f(\xbatch) \big) - \rho_u \noise_u \right\|^2 \nonumber\\
&\le 2 \E\|\sum_{u=o}^{B_\e-1} \eta_u \big( \nabla f_{i_{u}}(\x_{u+1}) - \nabla f_{i_u} (\xbatch) + \nabla f(\xbatch) \big)\|^2 + 2\E \|\sum_{u=o}^{B_\e-1} \rho_u \noise_u\|^2 \nonumber\\
&=2 \E\|\sum_{u=o}^{B_\e-1} \eta_u \big( \nabla f_{i_{u}}(\x_{u+1}) - \nabla f_{i_u} (\xbatch) + \nabla f(\xbatch) \big)\|^2 + 2\E \|\sum_{u=o}^{B_\e-1} \rho_u \noise_u\|^2 \nonumber\\
&\le 2 \E \sum_{u=o}^{B_\e-1} \eta_u^2 \left( \|\nabla f_{i_u}(\x_{u+1})\|^2 + \|\nabla f(\xbatch) - \nabla f_{i_u}(\xbatch)\|^2 \right) + 2d\sum_{u=o}^{B_\e-1} \rho_u^2 \nonumber\\
&\stackrel{\eqref{eqn::subset_selection_variance}}{\le} 2 D_F \sum_{u=o}^{B_\e-1} \eta_u^2 (1 + \frac{2}{B_\e})  + 2d\sum_{u=o}^{B_\e-1}\rho_u^2
\label{ineqn::batch_x_variation}
\end{align}
For function value decrease in the descent process, we have 
\begin{align*}
    f(\x_t) - \E f(\x_{t+1}) &\ge \E\left[ \<\nabla f(\x_t), \x_t - \x_{t+1} \> - \frac{L}{2}\|\x_t-\x_{t+1}\|^2 \right] \\
    &= \E\left[\<\nabla f(\x_t), \eta_k \gradsvrg_k \>  - \frac{L}{2}\| \eta_k \gradsvrg_k -\rho_k \noise_k\|^2\right]\\
    &= \E\left[\eta_t \|\nabla f(\x_t)\|^2  - \frac{L}{2}(\eta_t^2 \|\gradsvrg_t\|^2  + \rho_t^2 \|\noise_t\|^2)\right]\\
    &\ge \E\left[\eta_t \|\nabla f(\x_t)\|^2  - \frac{L}{2}\left(\eta_t^2 \big( 2 [\|\nabla f(\x_t)\|^2] + 2\frac{L^2}{B_\e}[\|\x_t-\xbatch\|^2] \big)  + \rho_t^2 \|\noise_t\|^2 \right)\right]\\
    &\stackrel{\eqref{ineqn::batch_x_variation}}{\ge} (\eta_t - \eta_t^2 L)\|\nabla f(\x_t)\|^2 - \underbrace{\frac{L^3}{B_\e} \eta_t^2 (3 D_F\sum_{u=o}^{B_\e-1}\eta_u^2 + 2d\sum_{u=o}^{B_\e-1}\rho_u^2) - \frac{L}{2}\rho_t^2 d}_{\mathcal{R}} = \mathcal{O}\left( \eps^2 \right)
\end{align*}
Here notice that $\sum_{u=o}^{B_\e-1} \eta_u^2 = \mathcal{O}(\eta_0\nu^{-1})$ and $\sum_{u=o}^{B_\e-1}\rho_u^2 = \mathcal{O}(\rho_0\nu^{-1})$, and set $B_\e = \max\{\frac{L^3D_f d}{\eps^2},1\}$ and $\delta = \mathcal{O}(r)$ (which consequently gives the order of $\eta_t^2$), then $\mathcal{R} = \mathcal{O}(\eps^2)$.

When a saddle point is encountered, within a minibatch with probability $(1-\frac{Q}{d T_i})P_i$, we have 
\begin{align*}
     &f(\x_o) - f(\x_o+\Delta_i) = f(\x_o) - f(\fsp) + f(\fsp) - f(\fsp + \Delta_i) + f(\fsp + \Delta_i) - f(\x_0 + \Delta_i) \\
     &= f(\fsp) - \left( f(\fsp) + \frac{1}{2}\Delta_i^\trans \nabla^2 f(\fsp)\Delta_i + \frac{L_2}{6}\|\Delta_i\|^3 \right) + f(\x_o) - f(\fsp) + f(\fsp + \Delta_i) - f(\x_0 + \Delta_i) \\
     &\ge f(\x_o) - f(\fsp) + f(\fsp + \Delta_i) - f(\x_0 + \Delta_i) - \frac{1}{2}\Delta_i^\trans \nabla^2 f(\fsp) \Delta_i + \frac{L_2}{6}\|\Delta_i\|^3 \\
     &\ge f(\x_o) - f(\fsp) + f(\fsp + \Delta_i) - f(\x_0 + \Delta_i) + \frac{\zeta}{2} - \frac{L_2}{6}r^3\\
     &\ge \frac{\zeta}{2} - \frac{L_2}{6}r^3 - 2Lr^2 = \mathcal{O}\left( \eps^2 \right)
\end{align*}
Set $\zeta = \frac{5\eps^2}{2 L} $, the time complexity to attain sufficient function value decrease before reaching a SSP is $\displaystyle \mathcal{O}\left(\frac{f(\x_0)-f_\star}{\eps^2} \right)$. 

\paragraph{Step 4} 
Now we give the description of $\tau_{\mathrm{SSP}}$ to finish the proof. In light of setting $\zeta = \frac{5}{2L}\eps^2$, consequently $r^2-Q = \frac{\eps^2}{2Lq(L+q)}$. From \eqref{defn::ssp_constraint_prob} together with \eqref{eqn::T_i_stepsize_batch}, the probability for constrained perturbation accumulation within the stepsize batch $i$ is given as $P_i = \mathcal{O}\left( \frac{\eps^{d-1}}{ \Gamma(\frac{d-2}{2}) L^{d-1}q^{d-1} } \right) \cdot \frac{1}{\exp(\mathcal{O}(i\delta d))}$. 

Assume the iteration sequence escapes saddle points in each stepsize batch where a saddle point is encountered. Setting the stepsize sum threshold $\delta = \mathcal{O}(\widetilde{\eps}/B)$ to be consistent with the setting of $\delta$ in the proof for theorem \ref{thm::ergodicity} and lemma \ref{lemma::reachability}, with probability $\mathcal{O}\left( \frac{\eps^{d-1}}{ \Gamma(\frac{d-2}{2}) L^{d-1}q^{d-1} } \right)$, the SGLD converges to a local minimum within time
\begin{equation}
    \tau_{\mathrm{SSP}} = \underbrace{\mathcal{O}\left(\frac{f(\x_0) - f_\star}{\eps^2}\right)}_{\textrm{descent}} + \underbrace{\exp\left(\mathcal{O}(\eps d)\right)}_{\textrm{escaping saddles}},
\end{equation}
where the first term accounts for the step needed for sufficient function value decrease, and the second term accounts for the time needed to escape saddle points as computed in equation \eqref{eqn::T_i_stepsize_batch}.

\end{proof}